\documentclass{article}

 \usepackage[main, final]{neurips_2026}

\usepackage[utf8]{inputenc} 
\usepackage[T1]{fontenc}    
\usepackage{hyperref}       
\usepackage{url}            
\usepackage{booktabs}       
\usepackage{amsfonts}       
\usepackage{nicefrac}       
\usepackage{microtype}      
\usepackage{xcolor}         

\usepackage{natbib}
\usepackage{booktabs} 
\usepackage{balance}
\usepackage{amsmath}
\usepackage{epsfig}
\usepackage{graphicx}
\usepackage{caption}
\usepackage{float}
\usepackage{subeqnarray}
\usepackage{cases}
\usepackage{amssymb}
\usepackage{amsmath}
\usepackage{amsthm}
\usepackage{bm}
\usepackage{url}
\usepackage{setspace}

\usepackage{multirow}
\usepackage{makecell}
\usepackage{threeparttable}
\usepackage{color}
\usepackage{makecell}

\theoremstyle{definition}

\usepackage{color}
\usepackage{bm}
\usepackage{amsmath}
\usepackage{stfloats}  
\usepackage{float}
\usepackage{caption}
\usepackage{multirow}
\usepackage{soul}

\usepackage{pifont}
\usepackage[most]{tcolorbox}

\definecolor{second}{HTML}{DAE8FC}
\definecolor{best}{HTML}{FFCCC9}
\newcommand{\secondhl}{\sethlcolor{second}\hl}
\newcommand{\besthl}{\sethlcolor{best}\hl}

\setcitestyle{numbers,square}
\usepackage{wrapfig}
\usepackage{algorithm}
\usepackage{algpseudocode}

\title{Hybrid Reinforcement Learning for\\ One-step Degraded Infrared and Visible Image Fusion}

\definecolor{linkblue}{HTML}{5B9BD5}
\hypersetup{
    colorlinks=true,
    allcolors=linkblue
}

\author{
\textbf{Hongwei Yu}$^{1,2,*}$ \quad
\textbf{Daoqing Zha}$^{1,*}$ \quad
\textbf{Jiawei Li}$^{1,\dagger}$ \quad
\textbf{Yancheng Bai}$^{2,\dagger,\ddagger}$ \\ 
\textbf{Lei Sun}$^{2}$ \quad
\textbf{Xiangxiang Chu}$^{2}$ \\[3pt]
$^{1}$University of Science and Technology Beijing \\
$^{2}$Amap, Alibaba Group \\
$^{*}$Equal contribution \qquad
$^{\dagger}$Corresponding authors \qquad
$^{\ddagger}$Project lead \\[4pt]
\url{https://yuhongwei22.github.io/onehrl-site/}
}

\begin{document}

\maketitle

\begin{abstract}
  Infrared and visible image fusion (IVIF) is essential for robust perception, yet existing methods remain limited under complex degradations.
  While diffusion models (DMs) offer powerful modeling capabilities to combat such degradations, their iterative process incurs high computational cost.
  One-step distillation effectively accelerates inference but typically leads to over-smoothed results.
  To resolve these limitations, we propose a one-step fusion framework with hybrid reinforcement learning, termed OneHRL.
  We treat degraded inputs as intermediate noisy states along a generative trajectory, enabling a distilled one-step generator to rectify them directly to the clean fusion results.
  Concurrently, to mitigate the over-smoothed results in distillation, we incorporate the Flow-GRPO paradigm to introduce stochasticity into the deterministic one-step generation.
  This restores the stochastic exploration necessary for sampling-based reinforcement learning.
  Furthermore, we design a coarse-to-fine reward that integrates a global semantic reward with an object-level reward.
  This hybrid reward ensures that the generator captures both naturalness and fine-grained texture.
  Extensive experiments demonstrate that our method achieves superior fusion quality and efficiency.
\end{abstract}

\section{Introduction}
Infrared and visible Image Fusion (IVIF) is a fundamental technique in multi-sensor perception, aiming to integrate the textural details of visible sensors with the thermal information of infrared ones \cite{li2026multi, liu2024infrared,liu2024promptfusion}.
While existing methods have demonstrated outstanding capabilities in ideal scenarios \cite{jiang2024multispectral, jiang2025harmonized, li2023gesenet}, real-world applications frequently suffer from various degradations.
Most previous works are designed for clear inputs and fail to generate robust outputs when facing complex degradations.
While some degradation-aware methods have emerged \cite{zhao2024image, bai2025refusion, yi2024text}, they are often restricted to single-type impairments and remain ineffective against composite degradations with complex distributions.

\begin{figure}
	\centering
	\includegraphics[width=0.8\linewidth]{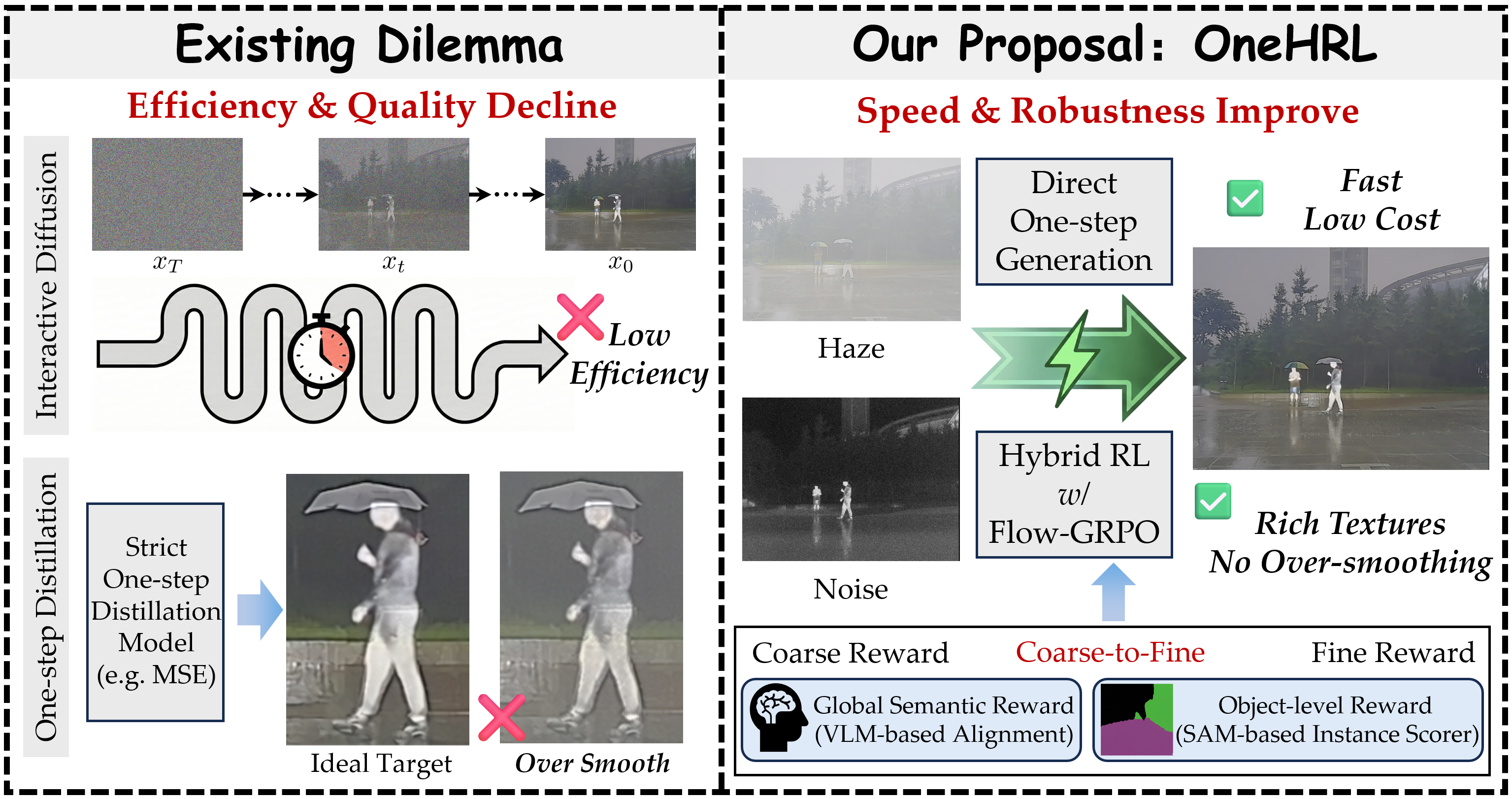}
	\caption{Schematic illustration of existing dilemma and our proposal. It is obvious that our proposed OneHRL can achieve both high inference speed and strong robustness.}
	\label{fig:first}
	\vspace{-15pt}
\end{figure}

Diffusion models (DMs) demonstrate powerful probabilistic modeling capabilities~\cite{li2025ldrps,lan2025flux}.
Recently, these capabilities of DMs have been used in IVIF to address degradation challenges.
Models such as DRMF \cite{tang2024drmf} and MMAIF \cite{Cao_2025_ICCV} have demonstrated remarkable robustness in degraded scenarios.
However, as shown in Fig.~\ref{fig:first}, the iterative sampling process of DMs introduces high computational costs, rendering them impractical for real applications.
To achieve acceleration, one-step distillation \cite{wu2024one} has been explored in tasks such as super-resolution (SR).
Nevertheless, its direct adaptation to the IVIF domain remains non-trivial.
IVIF necessitates the simultaneous alignment and fusion of disparate modalities, which increases task complexity compared with SR.

Meanwhile, strict one-step distillation (e.g., MSE) typically induces over-smoothed results \cite{he2022revisiting, dong2025tsd}.
To overcome this limitation, Sampling-based reinforcement learning (RL) \cite{shao2024deepseekmath,xu2025kdrl} is introduced to enhance textural richness, leveraging high-level perceptual or physical rewards.

Nevertheless, the deterministic nature of one-step generation poses a significant obstacle.
Specifically, since one-step distillation replaces random noise with a fixed degraded representation, the model performs a deterministic mapping.
This mapping lacks stochasticity, thereby precluding the exploration required for effective sampling-based RL optimization.

\begin{figure}
	\centering
	\includegraphics[width=0.98\linewidth]{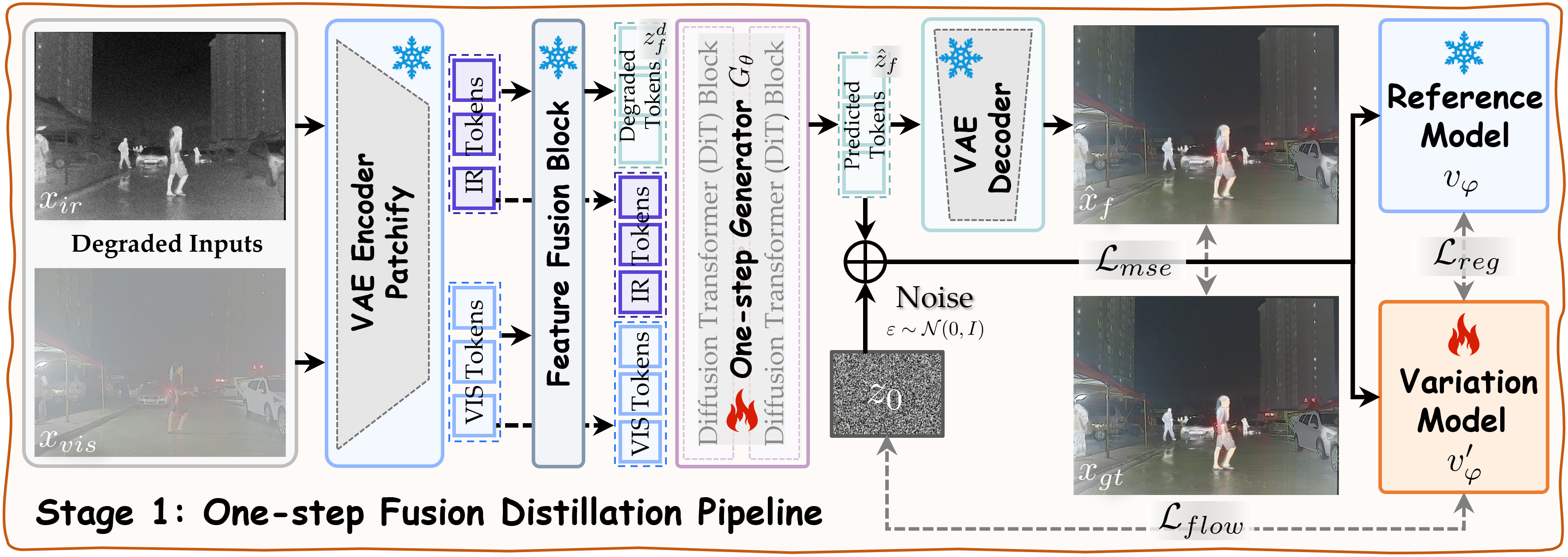}
	\caption{Framework of the proposed OneHRL. \textbf{One-step Distillation Stage:} Degraded inputs are mapped to the latent space and integrated into the degraded fused representation $z_f^d$ via the Feature Fusion Block. The generator $G_\theta$ then restores $z_f^d$ to a clean state in a single forward pass, optimized by $\mathcal{L}_{mse}$ and $\mathcal{L}_{reg}$. $\mathcal{L}_{flow}$ is utilized to update the variation model.}
	\label{fig:main1}
	\vspace{-15pt}
\end{figure}

\begin{figure}
	\centering
	\includegraphics[width=0.8\linewidth]{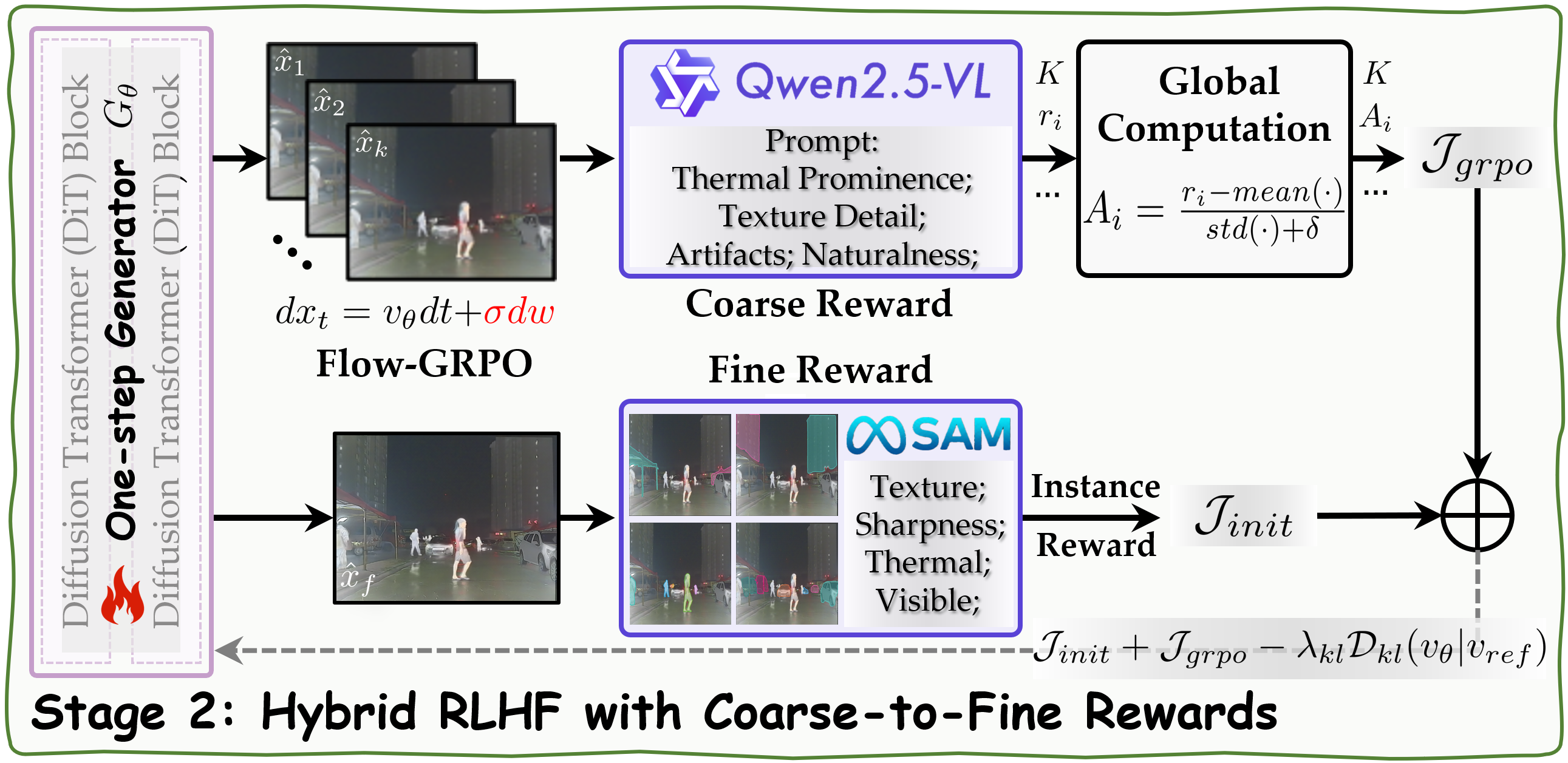}
	\caption{Framework of the proposed OneHRL. \textbf{Hybrid RL Stage:} The generator is refined via the Hybrid RL, leveraging a coarse-to-fine reward to enhance textural details.}
	\label{fig:main2}
	\vspace{-15pt}
\end{figure}

To address these challenges, we propose a novel one-step hybrid reinforcement learning fusion framework, termed OneHRL. The one-step and RL stages are respectively shown in Fig.~\ref{fig:main1} and \ref{fig:main2}.
Leveraging the concepts of generalized diffusion \cite{bansal2023cold}, we treat the degraded VIS/IR fusion process as a map between the corrupted and clean data manifolds.
Specifically, we process the degraded VIS/IR pairs through a feature fusion module pre-trained on clean data.
We then conceptualize the resulting representation as an intermediate noisy state and perform one-step distillation from a multi-step diffusion model.
It enables the generator to rectify this state directly toward the clean fusion manifold in a single step.
This mechanism leverages the feature fusion module for cross-modal information aggregation while exploiting the powerful distribution modeling capabilities of diffusion models to restore fidelity from complex degradations.
To mitigate over-smoothing, we incorporate the Flow-GRPO \cite{liu2025flow} paradigm, which reformulates the probability flow ordinary differential equation (ODE) into a stochastic differential equation (SDE) during the training phase. 
This transformation injects controlled stochasticity into the generative process, restoring the exploration space necessary for sampling-based RL.
Finally, we design a coarse-to-fine reward to provide multi-level supervision.
For the coarse reward, we leverage the large-scale vision-language model of Qwen2.5-VL \cite{bai2025qwen2} for the global semantic reward.
A differentiable instance scorer is designed to derive object-level rewards by leveraging the Segment Anything Model (SAM).
This hybrid reward enforces both global semantic coherence and object-level structural fidelity, ensuring robust fusion performance even under composite degradations. Our contributions are summarized as follows:
\begin{itemize}
	\item We propose an efficient one-step IVIF framework, which conceptualizes degraded inputs as intermediate noisy states. It enables the model to achieve joint restoration and modality fusion in a single step.
	\item We incorporate the Flow-GRPO paradigm into the one-step pipeline. By reformulating the probability flow ODE into an SDE, we successfully bridge the gap between deterministic process and sampling-based RL. It restores the stochastic exploration necessary.
	\item We design a hybrid coarse-to-fine reward that integrates the semantic intelligence of VLMs with a differentiable instance scorer based on SAM. It ensures the generator captures both global naturalness and object-level prominence, even under complex composite degradations.
	\item Extensive experiments conducted across multiple datasets and degradation scenarios validate that our OneHRL can achieve robust fusion results and maintain high accuracy in downstream tasks.
\end{itemize}

\section{Related Work and Preliminaries}
\subsection{Related Works}
\paragraph{Infrared-Visible Image Fusion for Degradation}
Most existing IVIF methods \cite{li2023learning, li2025a2rnet, ma2025bilevel, li2025difiisr, liang2025fusioninv, wang2026general, yang2026customized, li2025umcfuse} are primarily optimized for clean inputs and struggle in real-world degradations.
For the IVIF task under degraded conditions, a robust framework must simultaneously extract salient cross-modal features and mitigate the impacts of environmental artifacts on the fusion performance.
The scarcity of degraded scenarios in current benchmarks \cite{jia2021llvip, tang2022image, liu2022target} necessitate the use of simulation techniques to model degradations.
\cite{tang2025controlfusion} constructed an imaging model based on Retinex theory and the atmospheric scattering principle, designed a fine-tuning restoration and fusion network to deal with degraded source images. \cite{zhang2025omnifuse} introduced a language-driven framework for flexible and controllable degradation-robust image fusion. Moreover, \cite{xu2025urfusion} utilized intrinsic content extraction, intrinsic content fusion, and appearance representation learning and assignment to conduct a unified robust fusion framework.

\paragraph{Reinforcement Learning}
Reinforcement Learning has proven effective in aligning Large Language Models (LLMs) with human intent \cite{shao2024deepseekmath, chen2026d2align,chu2026gpg}.
While its initial success was prominent in aligning large language models with human preferences, recent advancements have extended RL to refine diffusion models for various vision tasks \cite{black2023training, fan2023dpok, kim2025chain,ma2026mar}.
The remarkable performance of RL in NLP and CV demonstrates its effectiveness in optimizing model performance.

\subsection{Preliminaries}
Recent advanced image-generation models \cite{bai2025qwen2} adopt Rectified Flow framework \cite{liu2022flow}, which connects two distributions via an Ordinary Differential Equation (ODE).
Let $z_0\sim \pi_0$ denote the prior distribution (e.g., standard Gaussian $\mathcal{N}(0, \mathbf{I})$) and $z_1\sim \pi_1$ denote the data distribution.
The dynamics are governed by the flow ODE:
\begin{equation}
	d z_t = v(z_t, t) dt,
\end{equation}
where $z_t$ is the state at time $t \in [0, 1]$, and $v$ is the velocity field. Ideally, the straight-line trajectory between a noise sample $z_0 \sim \pi_0$ and a data sample $z_1 \sim \pi_1$ is defined as:
\begin{equation}
	z_t = t z_1 + (1-t) z_0, \quad \text{with } v(z_t, t) = z_1 - z_0.
\end{equation}
In practice, we train a neural network $v_\theta$ to approximate this velocity field. The optimal parameters are obtained by minimizing the flow matching objective:
\begin{equation}
	\theta^* = \arg \min_{\theta} \mathbb{E} \left[ \left\| (z_1 - z_0) - v_\theta(t z_1 + (1-t) z_0, t) \right\|^2 \right].
\end{equation}

%

\section{Proposed Method}
\subsection{Motivation}
Existing advanced fusion methods demonstrate outstanding capabilities leveraging the DMs.
Despite their remarkable performance, the inherent nature of iterative sampling remains computationally expensive.
In SR tasks, recent works apply one-step distillation to compress the generation process.
Due to the distinct multimodal inputs of IVIF, directly applying such methods is challenging.
A suitable pipeline to achieve one-step fusion is lacking.
Moreover, strict distillation (e.g., MSE) often yields over-smoothed results.
Sampling-based RL (e.g., GRPO) has demonstrated the ability to improve perceptual quality and structural sharpness.
However, applying RL to one-step fusion models is non-trivial.
Since the inputs of one-step model are fixed, the outputs lack stochasticity, thereby precluding the exploration required by sampling-based RL methods.
To address these issues, we propose a one-step fusion framework with hybrid RL (OneHRL).
It conceptualizes degraded inputs as intermediate noisy states for direct restoration.
Furthermore, to enhance perceptual performance via RL, we propose the Hybrid RL strategy with Coarse-to-Fine Rewards.
For coarse semantic rewards, we introduce Flow-GRPO, a method that changes the ODE to an SDE while strictly guaranteeing the equivalence of the marginal distributions.
This transformation enables us to inject controlled stochasticity for sampling, which is then evaluated by Qwen2.5-VL to ensure overall naturalness.
For fine-grained rewards, we leverage the SAM to extract object instances and design a Instance Scorer to enforce pixel-level details.
This enables us to simultaneously guarantee efficiency and robustness.


\subsection{One-step Fusion via Distilled Diffusion Priors}
\label{sec:onestep}
To overcome the time cost while retaining the capability of diffusion models, we propose a one-step fusion framework based on rectified flow. Our goal is to learn a deterministic generator $G_\theta$ that directly maps degraded inputs to a clean, high-quality fusion image in a single forward pass.

\paragraph{Overview of Distillation Pipeline}
For notational brevity, we denote the input image space as $I$ and the compressed latent space encoded by the VAE as $z = \mathcal{E}(I)$.
We first train a feature fusion block (FFB), denoted as $\mathcal{F}_\phi$, using clean visible ($I_{vis}^c$) and infrared ($I_{ir}^c$) pairs. This module serves as a feature aggregator, optimized to maximize the retention of texture and radiation details:
\begin{equation}
	z_f^{c} = \mathcal{F}_\phi(z_{vis}^c, z_{ir}^c),
\end{equation}
where $z_{vis}^c, z_{ir}^c = \mathcal{E}(I_{vis}^c), \mathcal{E}(I_{ir}^c)$.
Simultaneously, we train a diffusion transformer (DiT) conditioned on degraded latent inputs $(z_{vis}^d, z_{ir}^d)$. This model functions as a restoration expert, learning a velocity field $v_\psi$ that governs the probability flow ODE from noise to data:
\begin{equation}
	d z_t = v_\psi(z_t, t, z_{vis}^d, z_{ir}^d) dt,\quad z_0 \sim \mathcal{N}(0, \mathbf{I}),
\end{equation}
where $z_t = t z_1 + (1-t) z_0$ and $z_1$ represents the encoded clean target $\mathcal{E}(x_{gt})$.
The pre-trained parameters of $\psi$ are used to initialize the one-step generator $G_\theta$, the frozen reference model $v_\psi$, and the learnable variational model $v_{\psi'}$.


In the distillation stage, we freeze the FFB and focus on distilling the restoration capability into a one-step generator. Given degraded inputs ($z_{vis}^d, z_{ir}^d$), the frozen FFB yields a degraded fused representation, denoted as $z_f^d$.
A key innovation of our method is to conceptualize $z_f^d$ not merely as a semantic condition, but as a specific intermediate noisy state along the generative trajectory.
Instead of sampling from pure Gaussian noise, we initialize the reverse process with $z_f^d$. 
The one-step generator then projects this state to the clean target manifold, followed by the VAE decoder $\mathcal{D}$ to obtain the final fused image $\hat{x}_f$:
\begin{equation}
	\begin{split}
		&\hat{x}_f = \mathcal{D}(\hat{z}_f) = \mathcal{D}( G_\theta(z_f^d, z_{vis}^d, z_{ir}^d) ), \\
		&G_\theta( \cdot) = z_f^d + \lambda_1v_{\theta}(z_f^d, z_{vis}^d, z_{ir}^d),
	\end{split}
\end{equation}
where $v_\theta$ represents the predicted velocity field, and $\lambda_1$ is a scaling factor (empirically set to $0.999$) controlling the rectification step size. $G_\theta$ thus learns a residual mapping to rectify the degraded trajectory in a single forward pass.

The generator is optimized via a composite objective. Specifically, $\hat{x}_f$ is constrained by a pixel-level reconstruction loss $\mathcal{L}_{mse}$ against the ground truth $x_{gt}$. Simultaneously, $\hat{z}_f$ is perturbed by noise and processed by both $v_\psi$ and $v_{\psi'}$ to derive the $\mathcal{L}_{reg}$. Furthermore, the variational model $v_{\psi'}$ is updated via a flow-matching objective $\mathcal{L}_{diff}$, minimizing the error between its output and the added noise to track the student distribution.

\paragraph{Loss of One-step Distillation}
To establish a deterministic one-step generator $G_\theta$, we employ a composite objective function. Our training strategy integrates a pixel-level reconstruction loss $\mathcal{L}_{mse}$ with a regularization term $\mathcal{L}_{reg}$ derived from diffusion priors.

We utilize the Mean Squared Error (MSE) between the generated output and the clean ground truth $x_f$ as the fundamental constraint:
\begin{equation}
	\mathcal{L}_{mse} = \| \mathcal{D}( G_\theta(z_f^d, z_{vis}^d, z_{ir}^d)) - x_{gt} \|^2_2.
\end{equation}
This term ensures the low-frequency structural fidelity of the fusion results, anchoring the content to the ground truth.
For regularization term, we employ Variational Score Distillation (VSD) as a regularizer to align the distribution of network outputs with the fusion results $\hat{x}_{f}$.
We instantiate two versions of the pre-trained DiT model: a frozen reference model $v_{\psi}$ and a learnable variational model $v_{\psi'}$.
The frozen model $v_{\psi}$ preserves the ground-truth velocity field learned from clean data. The variational model $v_{\psi'}$ acts as a dynamic teacher, tracking the student's current flow distribution,
\begin{equation}
	\begin{split}
		\mathcal{L}_{reg} &= \mathbb{E}_{t, z_0} \left[ || w(t) \left( v_{\psi'}(z_t, t, c) - v_{\psi}(z_t, t, c) \right) ||_2^2 \right], 
	\end{split}
\end{equation}
where $z_t = t\hat{z}_f + (1-t)z_0$, $\hat{z}_f = G_\theta(z_f^d, z_{vis}^d, z_{ir}^d)$, and $z_0\sim\mathcal{N}(0, \mathbf{I})$.
The total optimization objective for the generator $G_\theta$ is defined as:
\begin{equation}
	\mathcal{L}_{total} = \mathcal{L}_{mse} + \lambda_{reg} \mathcal{L}_{reg},
\end{equation}
where $\lambda_{reg}$ is a weighting scalar.
Furthermore, to ensure $v_{\psi'}$ effectively tracks the student's distribution, it is simultaneously updated via a flow matching objective $\mathcal{L}_{diff}$:
\begin{equation}
	\mathcal{L}_{diff} = \mathbb{E}_{t, z_0} \left[ \| v_{\psi'}(z_t, t, c) - (\hat{z}_f - z_0) \|^2_2 \right].
\end{equation}

		
		

\subsection{Hybrid RLHF with Coarse-to-Fine Rewards}
In the RL stage, we propose a Hybrid RLHF framework driven by a Coarse-to-Fine Reward.
We integrate a Flow-GRPO optimizer driven by VLMs for semantic alignment, and a differentiable instance scorer based on SAM for object-level precision.

\paragraph{Coarse RL via Flow-GRPO}
In the one-step generation, the stochastic input $z_t$ is replaced by the fixed input $z_f^d$. Consequently, the generator outputs a deterministic result $\hat{x}_f$.
It precludes the stochastic exploration essential for Sampling-based Reinforcement Learning (RL).
To address this, we introduce Flow-GRPO, which reformulates the ODE into an SDE.
For brevity, we denote the predicted velocity as $v_{\theta}(\cdot) = v_{\theta}(z_f^d,z_{vis}^d,z_{ir}^d)$. The SDE process is reformulated as:
\begin{equation}
	d z_t = \left[ v_{\theta}(\cdot) + \frac{\sigma_t^2}{2t} \left( z_f^d + (1-t)v_{\theta}(\cdot) \right) \right] dt + \sigma_t d w,
\end{equation}
where $dw$ denotes the Wiener process increments and $\sigma_t$ controls the level of stochasticity during generation.
This transformation injects controlled noise for exploration while theoretically preserving the marginal distributions of the generated samples.

For each degraded input, we sample a group of $K$ candidate outputs $\{ \hat{z}_1, \dots, \hat{z}_K \}$, creating a local distribution of potential fusion results.
We utilize Qwen2.5-VL as a semantic scorer $\mathcal{R}_{sem}$. To ensure robust evaluation, we design a Chain-of-Thought (CoT) prompt that instructs the VLM to analyze \textbf{Thermal Prominence}, \textbf{Texture Fidelity}, and \textbf{Overall Naturalness} before assigning a scalar score $r_i$.
The detailed prompt is shown in Appendix.
We then compute the relative advantage $A_i$ for each candidate based on the group statistics:
\begin{equation}
	A_i = \frac{r_i - \text{mean}(\{r_1, \dots, r_K\})}{\text{std}(\{r_1, \dots, r_K\}) + \delta},
\end{equation}
where $\delta$ is a constant for numerical stability.
The generator is updated to maximize the expected advantage. We employ the clipped surrogate objective to ensure stable policy updates:
\begin{small}
	\begin{equation}
		\mathcal{J}_{GRPO} = \mathbb{E} \left[ \frac{1}{k} \sum_i^k \left( \min\left( \rho_i A_i, \text{clip}(\rho_i, 1-\epsilon, 1+\epsilon) A_i \right) \right) \right],
	\end{equation}
\end{small}

where $\rho_i = \frac{p_\theta(\hat{z}_i|z_f^d,z_{vis}^d,z_{ir}^d)}{p_{\theta_{old}}(\hat{z}_i|z_f^d,z_{vis}^d,z_{ir}^d)}$ denotes the probability ratio between the current and old policies.

\paragraph{Fine RL via Instance Scorer}
While VLMs offer superior semantic guidance, they may overlook pixel-level distortions. To complement the coarse RL, we introduce an instance scorer that leverages the SAM for fine-grained RL.

For the fused image $\hat{x}_f$, we utilize a pre-trained SAM to extract a set of binary instance masks $M = \{m_1, \dots, m_N\}$.
Detailed extraction method can be found in Appendix.
This allows us to transition the evaluation from the global image level to the object level, focusing on regions of interest (e.g., pedestrians, vehicles).
For each extracted instance $m_j$, we construct a composite score based on five differentiable physical dimensions:
\textbf{Thermal Prominence ($\mathcal{S}_{tp}$)} evaluates saliency via Weber contrast and SNR,
\textbf{Local Sharpness ($\mathcal{S}_{ls}$)} penalizes blurring using Laplacian gradient energy,
\textbf{Information Entropy ($\mathcal{S}_{ie}$)} measures information richness via Shannon entropy,
\textbf{Infrared Visibility ($\mathcal{S}_{iv}$)} assesses ir preservation through intensity correlation,
and \textbf{Structural Similarity ($\mathcal{S}_{ssim}$)} enforces consistency with source modalities.
All operators are computed strictly within the instance masks.
The comprehensive formulations of these scores are provided in Appendix.
The instance-aware objective is formulated as the mean of these five metrics $\mathcal{K}:=\{tp, ls, ie, iv, ssim\}$:
\begin{equation}
	\mathcal{J}_{inst} = \frac{1}{N} \sum_{j=1}^{N} \sum_{k \in \mathcal{K}} \mathcal{S}_k(m_j),
\end{equation}
where $N$ denotes the total number of extracted instances.

\paragraph{Overall RL Optimization Objective}
The final optimization objective acts as a unified framework that balances semantic alignment, physical fidelity, and training stability. We formulate the total objective to be maximized as:
\begin{equation}
	\mathcal{J}_{total} = \mathcal{J}_{GRPO} + \lambda_{inst} \mathcal{J}_{inst} - \lambda_{kl} \mathcal{D}_{KL},
\end{equation}
where $\lambda_{inst}$ and $\lambda_{kl}$ are weighting hyperparameters.
Here, $\mathcal{D}_{KL}$ represents the Kullback-Leibler (KL) divergence penalty, which we approximate via the mean squared error between the velocity predictions of the current policy $v_\theta$ and the frozen reference model $v_{ref}$:
\begin{equation}
	\mathcal{D}_{KL} = \mathbb{E} \left[ \| v_\theta(z_t, c) - v_{ref}(z_t, c) \|^2_2 \right].
\end{equation}
This term serves as a crucial regularizer, preventing the generator from deviating excessively from the learned clean data manifold.

\section{Experiments}
\subsection{Experiment Setting}
\subsubsection{Degradation Scenario Construction}

To evaluate model robustness in degradation scenarios, we follow the simulation strategy of \cite{yi2024text} to synthesize challenging scenarios by applying diverse degradation conditions to clean source images.
All samples are from the $\text{M}^3\text{FD}$ and MSRS datasets.
The comprehensive training details and the degradation setting are in  Appendix.


\subsubsection{Comparison Methods and Metrics}
We benchmark our framework against seven SOTA methods for degradation scenario modeling, $i.e.$, ControlFusion \cite{tang2025controlfusion}, TG-ECNet \cite{sun2025task}, OmniFuse \cite{zhang2025omnifuse}, URFusion \cite{xu2025urfusion}, SDS-Fusion \cite{chen2025sdsfusion}, Text-IF \cite{yi2024text} and DRMF \cite{tang2024drmf}.
Furthermore, the downstream utility of the fused results is evaluated through object detection and semantic segmentation tasks using YOLOv5 \cite{redmon2016you} and DeepLabV3+ \cite{chen2018encoder}, measured by mean average precision (mAP@0.5) and mean intersection over union (mIoU), respectively.
For a more equitable comparison, we employ correlation coefficient (CC), the sum of the correlations of differences (SCD), mutual information (MI), visual information fidelity (VIF) and mean squared error (MSE) as metrics for quantitative assessment.


\subsection{Comparative Results}
\subsubsection{Fusion Results Comparison}

\begin{figure*}[t]
	\centering
	\includegraphics[width=0.98\linewidth]{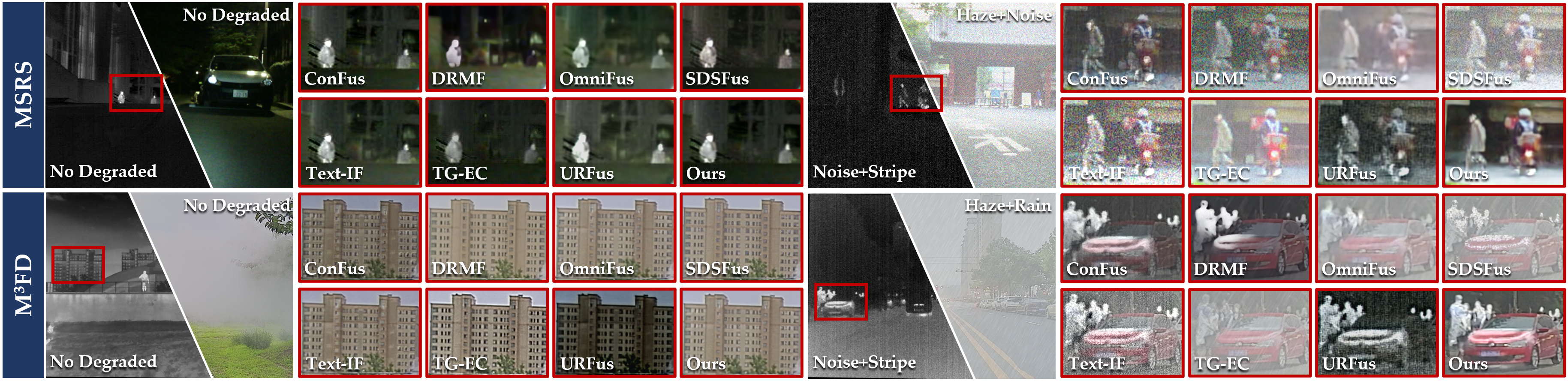}
	\caption{Fusion qualitative comparisons between SOTA methods and our OneHRL. Under different degradation conditions, the proposed OneHRL effectively mitigates adverse artifacts, which exhibit high visual quality.}
	\label{fig:fusion}	
    \vspace{-12pt}
\end{figure*}

\begin{table}[t]
	\centering
	\begin{minipage}{0.49\textwidth}
		\centering
		\caption{Fusion quantitative comparisons on the $\text{M}^3$FD dataset. Haze and rain degradations are added to VIS, while noise and stripe noise are incorporated into IR.}
		\resizebox{\linewidth}{!}{
			\begin{tabular}{c|ccccc}
				\toprule
				\multirow{2}{*}{\textbf{Method}} & \multicolumn{5}{c}{\textbf{Metric}} \\
				& CC$_{\scriptscriptstyle \uparrow}$ & SCD$_{\scriptscriptstyle \uparrow}$ & MI$_{\scriptscriptstyle \uparrow}$ & VIF$_{\scriptscriptstyle \uparrow}$ & MSE$_{\scriptscriptstyle \downarrow}$ \\
				\midrule
				ConFus & \secondhl{\textbf{0.492}} & \secondhl{\textbf{1.337}} & 2.363 & \secondhl{\textbf{0.487}} & 0.049 \\
				DRMF & 0.453 & 1.304 & 1.091 & 0.225 & 0.048 \\
				OmniFus & 0.418 & 0.779 & 2.335 & 0.386 & 0.112 \\
				SDSFus & 0.311 & 0.846 & 0.730 & 0.216 & 0.048 \\
				Text-IF & 0.465 & 0.833 & 2.156 & 0.443 & 0.118 \\
				TG-EC & 0.432 & 1.059 & \secondhl{\textbf{2.753}} & 0.454 & \secondhl{\textbf{0.044}} \\
				URFus & 0.416 & 0.810 & 2.497 & 0.420 & 0.117 \\
				Ours & \besthl{\textbf{0.519}} & \besthl{\textbf{1.544}} & \besthl{\textbf{3.088}} & \besthl{\textbf{0.498}} & \besthl{\textbf{0.039}} \\
				\bottomrule
			\end{tabular}
		}
		\label{tab:fusion_M3FD}
	\end{minipage}
	\hfill
	\begin{minipage}{0.49\textwidth}
		\centering
		\caption{Fusion quantitative comparisons on the MSRS dataset. Haze and noise degradations are added to VIS, while noise and stripe noise are incorporated into IR.}
		\resizebox{\linewidth}{!}{
			\begin{tabular}{c|ccccc}
				\toprule
				\multirow{2}{*}{\textbf{Method}} & \multicolumn{5}{c}{\textbf{Metric}} \\
				& CC$_{\scriptscriptstyle \uparrow}$ & SCD$_{\scriptscriptstyle \uparrow}$ & MI$_{\scriptscriptstyle \uparrow}$ & VIF$_{\scriptscriptstyle \uparrow}$ & MSE$_{\scriptscriptstyle \downarrow}$ \\
				\midrule
				ConFus & 0.562 & \secondhl{\textbf{0.920}} & 1.569 & \secondhl{\textbf{0.351}} & 0.126 \\
				DRMF & 0.558 & 1.021 & 1.137 & 0.320 & \secondhl{\textbf{0.049}} \\
				OmniFus & 0.552 & 0.734 & \secondhl{\textbf{1.577}} & 0.342 & 0.182 \\
				SDSFus & 0.435 & 0.752 & 0.861 & 0.243 & 0.149 \\
				Text-IF & \secondhl{\textbf{0.568}} & 0.824 & 1.387 & 0.333 & 0.219 \\
				TG-EC & 0.443 & 0.740 & 0.825 & 0.234 & 0.097 \\
				URFus & 0.429 & 0.648 & 0.906 & 0.258 & 0.222 \\
				Ours & \besthl{\textbf{0.611}} & \besthl{\textbf{1.527}} & \besthl{\textbf{2.384}} & \besthl{\textbf{0.465}} & \besthl{\textbf{0.033}} \\
				\bottomrule
			\end{tabular}
		}
		\label{tab:fusion_MSRS}
	\end{minipage}
\end{table}

Fig.~\ref{fig:fusion} illustrates the qualitative comparison of our OneHRL against other methods.
In clean scenarios of the MSRS dataset, DRMF and URFusion exhibit over-smoothed and a loss of details.
OmniFuse and TG-ECNet fail to preserve background textures.
ControlFusion, SDSFusion and Text-IF introduce unnatural color casts.
Our OneHRL maintains superior structural fidelity and color consistency.
Under composite degradations, SDSFusion, DRMF and Text-IF can alleviate haze but leave the noise influence.
Conversely, OmniFuse and TG-ECNet suppress noise but are ineffective against haze interference.
ControlFusion and URFusion remain degraded in the IR modality.
In contrast, our proposed OneHRL effectively eliminates diverse degradations while maintaining rich-detailed textures and structural fidelity.
Similarly, on the $\text{M}^3\text{FD}$ dataset, OneHRL achieves exceptional fusion performance under clean inputs and maintains robustness across diverse degraded scenarios.
Tab.~\ref{tab:fusion_M3FD} summarizes the quantitative comparisons under the composite degradation of Haze+Rain (VIS) and Noise+Stripe (IR) on the $\text{M}^3\text{FD}$ dataset. Similarly, Tab.~\ref{tab:fusion_MSRS} presents the quantitative results for the MSRS dataset under Haze+Noise (VIS) and Noise+Stripe (IR) conditions.
It demonstrates that OneHRL consistently achieves top-tier performance across all metrics.
These results validate OneHRL's robustness and yields exceptional fused images across diverse degradation scenarios.
Additional evaluations, including results under clean conditions, are provided in Appendix.

\begin{figure*}[t]
	\centering
	\includegraphics[width=0.98\linewidth]{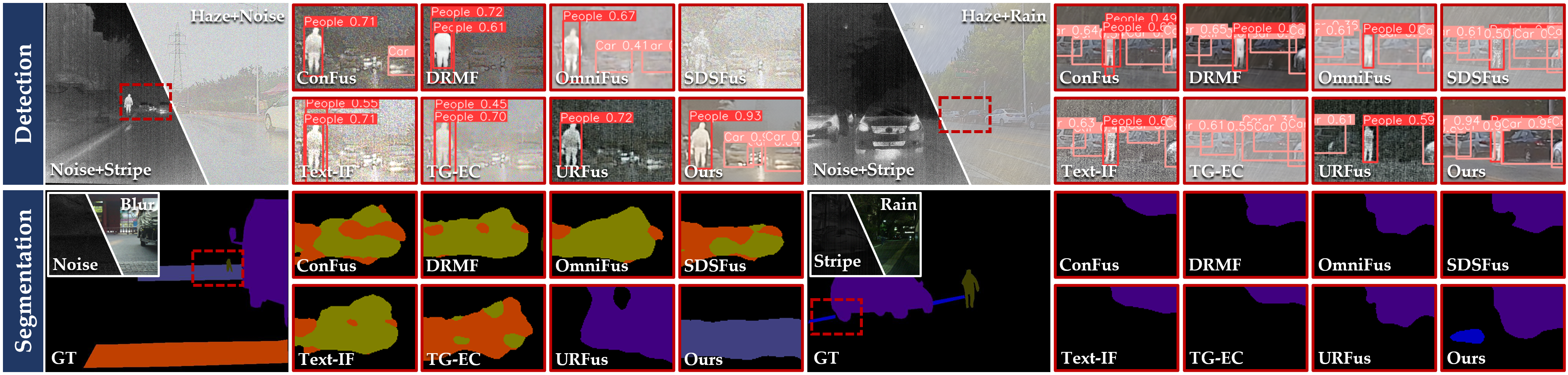}
	\caption{Downstream qualitative comparisons between SOTA methods and our OneHRL. Under different degradation conditions, the proposed OneHRL effectively mitigates adverse artifacts, which can exhibit both high visual quality and reliable downstream performance.}
	\label{fig:downstream}	
\end{figure*}

\subsubsection{Detection Results Comparison}

To further evaluate the restoration capability of OneHRL, we display detection comparisons of different degradation in Fig.~\ref{fig:downstream}.
Under the composite degradation of haze and noise, OneHRL successfully localizes all targets, including one person and three cars.
In contrast, DRMF, Text-IF, and TG-ECNet fail to detect the vehicle targets and produce redundant boxes for the person.
ControlFusion, URFusion, and OmniFuse correctly identify the pedestrian, but missing vehicle detections.
SDSFusion fails to detect any targets.
Similarly, in scenarios of haze and rain, ControlFusion, SDSFusion, and TG-ECNet produce false positives.
DRMF, OmniFuse, and Text-IF locate error regions.
URFusion further suffers from target missing.
Tab.~\ref{down-stream_quantitative} summarizes the quantitative evaluation in terms of mAP@.5.
Compared with other SOTA methods, OneHRL achieves a competitive mAP of 47.0\% under the joint interference of Noise+Stripe (IR) and Haze+Noise (VIS), and 56.8\% in the Noise+Stripe (IR) and Haze+Rain (VIS) scenario.
These results demonstrate that our OneHRL yields high-quality results which are suitable for the detection task.

\subsubsection{Segmentation Results Comparison}

As illustrated in Fig.~\ref{fig:downstream}, we present two sets of qualitative segmentation comparisons to evaluate the semantic consistency of our fused results.
In the left example, it contains a guardrail and a pedestrian.
The URFusion misclassifies the guardrail as a vehicle under the degradation interference.
Other methods produce segmentations consisting of bump and car stop categories, and exhibit region misalignments relative to the ground truth.
The right one is collected from a dark environment.
Other methods fail to fit the vehicle contours and completely neglect the road curve.
Our OneHRL successfully segments the road curve and achieves more precise alignment with the vehicle boundaries.
The quantitative mIoU results for each method are summarized in Tab.~\ref{down-stream_quantitative}. Taken together, both qualitative and quantitative analyses confirm that OneHRL maintains high segmentation accuracy despite complex degradations.

\begin{table*}[t]
	\centering
	\caption{Quantitative comparisons of downstream tasks. The subscripts represent the performance drop compared with the corresponding clean results. N-S, H-N, and H-R represent combined degradations involving Noise+Stripe, Haze+Noise, and Haze+Rain, respectively. The best and second-best results are highlighted in \besthl{red} and \secondhl{blue}.}
	\label{down-stream_quantitative}
	
	\setlength{\tabcolsep}{2pt}
	\renewcommand{\arraystretch}{1.08}
	
	\resizebox{\textwidth}{!}{
		\begin{tabular}{c|c|cccccccc}
			\toprule
			\multirow{2}{*}{\textbf{Degradation Type}} &
			\multirow{2}{*}{\textbf{Metric}} &
			\multicolumn{8}{c}{\textbf{Method}} \\
			& & ConFus & DRMF & OmniFus & SDSFus & Text-IF & TG-EC & URFus & Ours \\
			\midrule
			
			$\text{N-S} + \text{H-N}$ & \multirow{2}{*}{$\mathrm{mAP@.5}$} &
			\secondhl{\textbf{0.196}}$_{\scriptscriptstyle \downarrow0.459}$ &
			0.165$_{\scriptscriptstyle \downarrow0.413}$ &
			0.160$_{\scriptscriptstyle \downarrow0.440}$ &
			0.081$_{\scriptscriptstyle \downarrow0.405}$ &
			0.156$_{\scriptscriptstyle \downarrow0.503}$ &
			0.134$_{\scriptscriptstyle \downarrow0.535}$ &
			0.172$_{\scriptscriptstyle \downarrow0.474}$ &
			\besthl{\textbf{0.470}}$_{\scriptscriptstyle \downarrow0.159}$ \\
			
			$\text{N-S} + \text{H-R}$ &  &
			\secondhl{\textbf{0.457}}$_{\scriptscriptstyle \downarrow0.198}$ &
			0.367$_{\scriptscriptstyle \downarrow0.211}$ &
			0.249$_{\scriptscriptstyle \downarrow0.351}$ &
			0.175$_{\scriptscriptstyle \downarrow0.311}$ &
			0.285$_{\scriptscriptstyle \downarrow0.374}$ &
			0.231$_{\scriptscriptstyle \downarrow0.438}$ &
			0.153$_{\scriptscriptstyle \downarrow0.493}$ &
			\besthl{\textbf{0.568}}$_{\scriptscriptstyle \downarrow0.061}$ \\
			
			\midrule
			
			$\text{Noise} + \text{Blur}$ & \multirow{2}{*}{$\mathrm{mIoU}$} &
			\secondhl{\textbf{0.527}}$_{\scriptscriptstyle \downarrow0.039}$ &
			0.521$_{\scriptscriptstyle \downarrow0.027}$ &
			0.496$_{\scriptscriptstyle \downarrow0.029}$ &
			0.407$_{\scriptscriptstyle \downarrow0.153}$ &
			0.513$_{\scriptscriptstyle \downarrow0.056}$ &
			0.508$_{\scriptscriptstyle \downarrow0.031}$ &
			0.275$_{\scriptscriptstyle \downarrow0.276}$ &
			\besthl{\textbf{0.601}}$_{\scriptscriptstyle \downarrow0.018}$ \\
			
			$\text{Stripe} + \text{Rain}$ &  &
			0.441$_{\scriptscriptstyle \downarrow0.125}$ &
			0.385$_{\scriptscriptstyle \downarrow0.164}$ &
			\secondhl{\textbf{0.449}}$_{\scriptscriptstyle \downarrow0.076}$ &
			0.374$_{\scriptscriptstyle \downarrow0.185}$ &
			0.390$_{\scriptscriptstyle \downarrow0.180}$ &
			0.352$_{\scriptscriptstyle \downarrow0.187}$ &
			0.401$_{\scriptscriptstyle \downarrow0.150}$ &
			\besthl{\textbf{0.609}}$_{\scriptscriptstyle \downarrow0.010}$ \\
			
			\bottomrule
		\end{tabular}
	}
\end{table*}

\subsection{Ablation Analysis}
\subsubsection{Analysis of Modules}

\begin{figure}[h]
	\centering
	
	\begin{minipage}[t]{0.49\linewidth}
		\vspace{2pt}
		\centering
		\includegraphics[width=\linewidth]{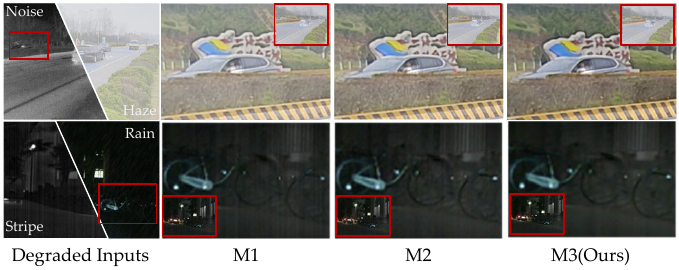}
		\captionof{figure}{Ablation qualitative analysis of modules. Two scenes i.e., noise with haze and stripe with haze are selected in the $\text{M}^3$FD and MSRS dataset.}
		\label{fig:ablation-module}	
	\end{minipage}
	\hfill
	\begin{minipage}[t]{0.49\linewidth}
		\centering
		\captionof{table}{Ablation quantitative analysis of modules. 
		}
		\renewcommand\tabcolsep{3.8pt}
		
		\resizebox{\linewidth}{!}{
			\begin{tabular}{cc|cc|ccccc}
				\toprule
				\multicolumn{2}{c|}{\multirow{2}{*}{\textbf{Model}}} & \multicolumn{2}{c|}{\textbf{Module}} & \multicolumn{5}{c}{\textbf{Metric}}                               \\
				\multicolumn{2}{c|}{} & $\mathcal{D}$ & $\mathcal{R}$ & CC & SCD & MI & VIF & MSE            
				\\ 
				\midrule
				\multirow{3}{*}{\rotatebox{90}{\textbf{$\text{M}^{\text{3}}$FD}}} & M1 & \ding{55} & \ding{55} & 0.431 & 1.395 & 2.876 & 0.511 & 0.128 
				\\
				& M2 & \ding{51} & \ding{55} & 0.462 & 1.477 & 3.009 & 0.534 & 0.086          
				\\
				& M3 & \ding{51} & \ding{51} & \besthl{\textbf{0.520}} & \besthl{\textbf{1.544}} & \besthl{\textbf{3.310}} & \besthl{\textbf{0.573}} & \besthl{\textbf{0.059}}          
				\\ 
				\midrule
				\multirow{3}{*}{\rotatebox{90}{\textbf{MSRS}}} & M1 & \ding{55} & \ding{55} & 0.573 & 1.445 & 2.316 & 0.555 & 0.046          
				\\
				& M2 & \ding{51} & \ding{55} & 0.591 & 1.537 & 2.764 & 0.632 & 0.051          
				\\
				& M3 & \ding{51} & \ding{51} & \besthl{\textbf{0.620}} & \besthl{\textbf{1.618}} & \besthl{\textbf{2.982}} & \besthl{\textbf{0.705}} & \besthl{\textbf{0.032}}          
				\\ 
				\bottomrule
			\end{tabular}
		}
		\label{quantitative ablation module}
	\end{minipage}
	\vspace{-5pt}
\end{figure}

To demonstrate the effectiveness of each proposed component, Fig.~\ref{fig:ablation-module} presents an ablation study on the $\text{M}^3$FD and MSRS datasets.
Here, $\mathcal{D}$ denotes the one-step distillation and $\mathcal{R}$ represents the RL training.
We employ a baseline DiT model (M1) with 5-step iterative sampling.
Subsequent models progressively incorporate these modules to form the final OneHRL.
Under the composite degradation of noise and haze, M2 (M1+D) enhances the robustness against degradations. It also effectively improves the inference speed.
By integrating the RL, OneHRL (M3) further refines textural details and visual clarity.
The other examples also show the consistent performance gains.
As shown in Tab.~\ref{quantitative ablation module}, the results confirm that OneHRL achieves the best performance across all fusion metrics.


\subsubsection{Analysis of Rewards}

\begin{figure}[h]
	\centering
	
	\begin{minipage}[t]{0.49\linewidth}
		\vspace{-5pt}
		\centering
		\includegraphics[width=\linewidth]{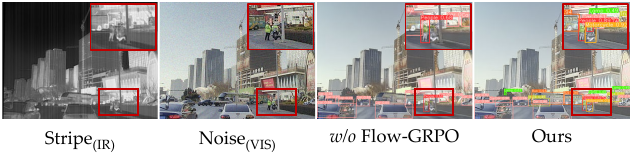}
		\captionof{figure}{Ablation qualitative analysis of rewards. We show their performance in the detection tasks.}
		\label{fig:ablation-RL}	
	\end{minipage}
	\hfill
	\begin{minipage}[t]{0.49\linewidth}
		\centering
		\captionof{table}{Ablation quantitative analysis of modules. 
		}
		\renewcommand\tabcolsep{3.8pt}
		
		\resizebox{\linewidth}{!}{
			\begin{tabular}{c|cccccc|c}
			\toprule
			\multirow{2}{*}{\textbf{Method}} & \multicolumn{6}{c|}{$\mathrm{AP@.5}$}                                                                           & \multirow{2}{*}{$\mathrm{mAP@.5}$} 
			\\
			& People & Car & Bus & Motor & Truck & Lamp &                        
			\\ 
			\midrule
			\textit{w/o} F-G & 0.516 & 0.763 & 0.429 & 0.117 & 0.492 & 0.244 & 0.426                   
			\\
			Ours & \besthl{\textbf{0.637}} & \besthl{\textbf{0.825}} & \besthl{\textbf{0.630}} & \besthl{\textbf{0.338}} & \besthl{\textbf{0.601}} & \besthl{\textbf{0.335}} & \besthl{\textbf{0.561}}          
			\\ 
			\bottomrule
		\end{tabular}
		}
		\label{Ablation RL}
	\end{minipage}
    \vspace{-5pt}
	
\end{figure}

In this section, we compare the performance of using only the instance reward (denoted as w/o F-G) against the full configuration on the detection task.
We introduce the degradations of Noise (VIS) and Stripe Noise (IR).
While employing only the instance reward can improve the contrast for some objects, the semantic style may change.
It poses challenges for downstream tasks.
As shown in Fig.~\ref{fig:ablation-RL}, results of w/o Flow-GRPO miss several detections of target instances.
Tab.~\ref{Ablation RL} reports the per class AP@0.5 and mAP@0.5, which observe that the proposed OneHRL stands a leading position.

\subsubsection{Analysis of KL Constraints}

\begin{wrapfigure}{r}{0.6\textwidth}
	\centering
	\vspace{-12pt}
	\captionof{table}{Ablation quantitative analysis of the KL constraint. The evaluation is conducted on the segmentation task.}
	\resizebox{\linewidth}{!}
	{
		\renewcommand\tabcolsep{1pt} 
		\begin{tabular}{c|ccccccccc|c}
			\toprule
			\multirow{2}{*}{\textbf{Method}} & \multicolumn{9}{c|}{$\mathrm{IoU}$}                                                                                                                      & \multirow{2}{*}{$\mathrm{mIoU}$} \\
			& Bac & Car & Per & Bik & Cur & C S & Gua & C C & Bum &  
			\\ 
			\midrule
			\textit{w/o} KL & 0.973 & 0.653 & 0.501 & 0.449 & 0.156 & 0.498 & 0.423 & 0.379 & 0.413 & 0.493 
			\\
			Ours & \besthl{\textbf{0.974}} & \besthl{\textbf{0.824}} & \besthl{\textbf{0.597}} & \besthl{\textbf{0.569}} & \besthl{\textbf{0.249}} & \besthl{\textbf{0.591}} & \besthl{\textbf{0.687}} & \besthl{\textbf{0.436}} & \besthl{\textbf{0.510}} & \besthl{\textbf{0.601}} 
			\\ 
			\bottomrule
		\end{tabular}
	}
	\label{ablation KL}
	\vspace{0.3em}
	
	\includegraphics[width=0.98\linewidth]{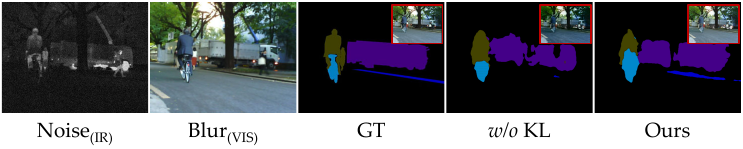}
	\captionof{figure}{Ablation qualitative analysis of the KL constraint. We show their performance in the segmentation task.}
	\label{fig:ablation-KL}	
	\vspace{-10pt}
\end{wrapfigure}

	

The KL constraint is a crucial regularization term in reinforcement learning that prevents the model from deviating excessively from the original one.
We evaluate its impact through an ablation study on the semantic segmentation task.
Fig.~\ref{fig:ablation-KL} illustrates the qualitative results under the interference of Blur (VIS) and Noise (IR).
Without the KL constraint, the model exhibits instability, which leads to inaccurate segmentation masks.
Specifically, distorted shapes appear in the vehicle regions, and the majority of the road curvature is lost.
In contrast, our full method provides precise segmentation results and successfully preserves all semantic elements.
In addition, IoU for each class and mIoU is shown in Tab.~\ref{ablation KL}.
OneHRL achieves superior performance across all categories.


\subsubsection{Analysis of Generation Speed}
In this section, we evaluate the inference efficiency of our proposed method compared to the diffusion-based baseline, DRMF.
While DRMF requires 25 seconds to process a single image due to its iterative sampling process, OneHRL completes the generation in just 0.4 seconds.
This represents a $62.5\times$ speedup, significantly enhancing its practical utility for the fusion task.

\section{Conclusion}
In this paper, we propose OneHRL, a one-step robust infrared and visible image fusion framework driven by hybrid reinforcement learning.
By integrating one-step distillation with a coarse-to-fine reward, OneHRL effectively addresses the speed-robustness dilemma inherent in traditional diffusion-based models.
Through the one-step distillation, our method achieves a significant speed improvement compared to existing diffusion-based approaches.
Furthermore, our approach utilizes Flow-GRPO to ensure global semantic consistency and an instance scorer to enhance local textural details.
Experimental results on the $\text{M}^3$FD and MSRS datasets demonstrate that OneHRL achieves both speed and robustness.
Under diverse degradations, our method exhibits superior anti-degradation performance and yields consistent effectiveness across various downstream tasks.

\section{Funding Transparency Statement}
This work was conducted during Hongwei Yu's internship at Alibaba Group. Alibaba Group provided financial support and computational resources for this work. The authors declare no other competing interests.

{
\small

\bibliographystyle{unsrtnat}
\bibliography{NIPS26_OneHRL}
}


\newpage

\appendix

\section*{Technical appendices and supplementary material}

\section{The Details of Our Algorithms}
\begin{algorithm}
\caption{One-step Distillation for Image Fusion}
\label{alg:distillation}
\begin{algorithmic}[1]
\Require Degraded pairs $(I_{vis}^d, I_{ir}^d)$, ground truth $x_{gt}$, VAE $\{\mathcal{E}, \mathcal{D}\}$, frozen FFB $\mathcal{F}_\phi$.
\Require Initialize $G_\theta, v_\psi, v_{\psi'}$ with pre-trained $\psi$.
\While{not converged}
\State \textit{// Step 1: Forward Pass}
\State $z_f^d = \mathcal{F}_\phi(\mathcal{E}(I_{vis}^d), \mathcal{E}(I_{ir}^d))$; {\Comment{\footnotesize Encode and fuse degraded inputs}}
\State $\hat{z}_f = G_\theta(z_f^d, \cdot)$; $\hat{x}_f = \mathcal{D}(\hat{z}_f)$; {\Comment{\footnotesize One-step prediction and reconstruction}}

\State \textit{// Step 2: Optimization Objectives}
\State Sample $t \sim \mathcal{U}(0,1), z_0 \sim \mathcal{N}(0, \mathbf{I})$; Compute $z_t = t\hat{z}_f + (1-t)z_0$;
\State $\mathcal{L}_{reg} = || w(t) (v_{\psi'} - v_{\psi}) ||_2^2$; $\mathcal{L}_{mse} = || \hat{x}_f - x_{gt} ||_2^2$;
\State $\mathcal{L}_{diff} = || v_{\psi'}(z_t, t, c) - (\hat{z}_f - z_0) ||^2_2$;

\State \textit{// Step 3: Parameter Update}
\State $\theta \leftarrow \nabla_\theta (\mathcal{L}_{mse} + \lambda_{reg} \mathcal{L}_{reg})$; {\Comment{\footnotesize Update generator}}
\State $\psi' \leftarrow \nabla_{\psi'} \mathcal{L}_{diff}$; {\Comment{\footnotesize Update variation model for tracking}}
\EndWhile
\State \textbf{return} Optimized generator $G_\theta$
\end{algorithmic}
\end{algorithm}

\begin{algorithm}
\caption{Hybrid RLHF Training for Robust Fusion}
\label{alg:rlhf}
\begin{algorithmic}[1]
\Require Generator $G_{\theta}$, reference $v_{ref}$, semantic scorer $\mathcal{R}_{sem}$, and instance extractor (SAM).
\While{not converged}
\State \textit{// Step 1: Flow-GRPO Group Sampling \& Advantage}
\For{$i = 1$ to $K$}
\State Sample $\hat{z}_i$ via SDE; Decode $\hat{x}_i = \mathcal{D}(\hat{z}_i)$;
\State $r_i = \mathcal{R}_{sem}(\hat{x}_i)$; {\Comment{\footnotesize VLM semantic scoring}}
\EndFor
\State $A_i = \frac{r_i - \text{mean}(\{r_k\})}{\text{std}(\{r_k\}) + \delta}$; {\Comment{\footnotesize Compute group-wise advantage}}

\State \textit{// Step 2: Instance Reward}
\State $\hat{x}_f = \mathcal{D}(G_{\theta}(z_f^d, \cdot))$; {\Comment{\footnotesize Primary fused result}}
\State $\mathcal{J}_{inst} = \text{SAM-Reward}(\hat{x}_f)$; {\Comment{\footnotesize Object-level fidelity}}

\State \textit{// Step 3: Combined Optimization}
\State {\small $\mathcal{J}_{total} = \mathcal{J}_{GRPO}(A_i) + \lambda_{inst} \mathcal{J}_{inst} - \lambda_{kl} \mathcal{D}_{KL}(v_\theta, v_{ref});$}
\State Update $\theta$ by maximizing $\mathcal{J}_{total}$;
\EndWhile
\State \textbf{return} Refined generator $G_{\theta}$
\end{algorithmic}
\end{algorithm}

Algorithms~\ref{alg:distillation} and~\ref{alg:rlhf} summarize the overall training pipeline of our framework. 
We first perform one-step distillation in the latent space, where the degraded visible and infrared inputs are encoded by the VAE and fused by the frozen feature fusion block. 
The generator $G_\theta$ then predicts the fused latent in a single forward pass, which is decoded to obtain the reconstructed fusion image. 
During this stage, the generator is optimized with the reconstruction loss to preserve pixel-level fidelity, while a regularized variation model is updated to track the target diffusion velocity. 
This distillation stage transfers the generation capability of the diffusion prior into an efficient one-step fusion generator.

After distillation, we further refine the generator through hybrid RLHF training. 
Specifically, Flow-GRPO samples a group of candidate fused results through the stochastic flow process and computes normalized group-wise advantages according to the semantic rewards produced by the VLM-based scorer. 
In parallel, an instance-level reward derived from SAM is introduced to encourage object-level structural fidelity. 
The final optimization objective combines the GRPO reward, the instance reward, and a KL regularization term with respect to the reference model, enabling the generator to improve semantic and instance-level robustness while maintaining stable generation behavior.

\section{Details of Experiment Setting}
\subsection{Degradation Scenario Construction}
To evaluate model robustness in degradation scenarios, we follow the simulation strategy of \cite{yi2024text} to synthesize challenging scenarios by applying diverse degradation conditions to clean source images.
Specifically, the degradation types include blur, haze, rain, noise, and stripe.
To ensure physical plausibility and prevent excessive information loss, each pristine image is subject to at most two concurrent degradations.
This strategy of stochastic composite degradations encourages the model to learn generalized restoration features rather than over-fitting to specific cases. All samples from the $\text{M}^3\text{FD}$ and MSRS datasets are processed following this protocol.

\subsection{Training Details}
The generator utilizes a DiT backbone consisting of 8 layers, where each layer features 8 attention heads with a head dimension of 128.
The input and output channels are configured as 64 and 16, respectively.
We employ a pre-trained VAE encoder from Qwen-Image-Edit \cite{wu2025qwen} for feature extraction.
During the training phase, the AdamW optimizer \cite{loshchilov2017decoupled} is used to update the parameters of the student network, with the weight decay and learning rate set to $1e^{-2}$ and $1e^{-4}$, respectively. 
Clean and degraded source images from the $\text{M}^3\text{FD}$ and MSRS datasets are utilized for training, which adopts a random pairing strategy for enhanced generalization.
In the distillation stage, we set the total number of steps to 50k.
The hyperparameters $\lambda_{reg}$ is set to $1e^{-2}$.
In the RL stage, we train the model for 1k steps.
The hyperparameters $\lambda_{inst}$ and $\lambda_{kl}$ are set to $1e^{-2}$ and $0.8$.
All experiments are performed on a PyTorch framework with 8 NVIDIA H20 Tensor Core GPUs and an Intel(R) Xeon(R) Platinum 8480+ CPU. Notably, the model maintains high efficiency, requiring only 5GB of memory during the inference stage.

\section{Technical Details of Instance Scorer}
\subsection{Instance Extraction via SAM }
To facilitate object-level evaluation in the reinforcement learning stage, we employ the Segment Anything Model (SAM) with a ViT-H backbone to partition the fused image $\hat{x}_f$ into instances. The extraction process is implemented via the SamAutomaticMaskGenerator, which identifies regions of interest without requiring specific category labels.

We configure the mask generator to prioritize both coverage and mask quality. The key parameters are set as follows:
\begin{itemize}
	\item Grid Sampling: We utilize a $32 \times 32$ point-grid ($1024$ points in total) sampled across the image to ensure that small yet salient objects (e.g., distant pedestrians) are not overlooked.
	\item  Confidence Filtering: To ensure the instance-level rewards are derived from high-quality masks, we set the predicted IoU threshold to 0.86 and the stability score threshold to 0.92. Masks failing to meet these criteria are discarded to avoid providing noisy feedback to the generator $G_\theta$.
	\item Area Constraint: We exclude fragmented regions with a total area of fewer than 100 pixels. This prevents the fine-grained RL stage from being distracted by sensor-level noise or trivial artifacts that do not represent physical instances.
\end{itemize}
For the instance-level evaluation, the Segment Anything Model (SAM) is employed to partition the primary fused output $\hat{x}_f$ into a collection of binary masks $M = \{m_1, \dots, m_N\}$. 
These masks identify semantically significant regions of interest, such as pedestrians or vehicles, allowing the reward system to transition from global image assessment to fine-grained object-level precision. Specifically, each extracted mask $m_j$ serves as a spatial operator to isolate instance-specific features for the computation of differentiable physical rewards, including thermal prominence ($\mathcal{S}_{tp}$) and local sharpness ($\mathcal{S}_{ls}$).

\subsection{Instance Rewards}
To provide fine-grained feedback for the reinforcement learning stage, we define a composite instance-level reward $\mathcal{J}_{inst}$ based on five differentiable physical dimensions. Each metric is computed strictly within the spatial support of the binary mask $m_j$ extracted by SAM.
For each extracted mask $m_j$, we define the localized instance representations as $\hat{x}_f^j = \hat{x}_f \odot m_j$, $x_{ir}^j = x_{ir} \odot m_j$, and $x_{vis}^j = x_{vis} \odot m_j$. The reward $\mathcal{J}_{inst}$ is then computed as a weighted sum of five differentiable metrics, each operating on these masked regions.

\paragraph{1. Thermal Prominence ($\mathcal{S}_{tp}$)}
This reward ensures infrared targets remain salient relative to their local surroundings. We define an inner target region $\mathcal{M}_{in}$ via morphological erosion and a background region $\mathcal{M}_{bg}$ via dilation of the mask $m_j$. The score combines three sub-metrics:
\begin{itemize}
	\item \textbf{Weber Contrast}: $C_{w} = \frac{\mu(\hat{x}_f^j | \mathcal{M}_{in}) - \mu(\hat{x}_f^j | \mathcal{M}_{bg})}{\mu(\hat{x}_f^j | \mathcal{M}_{bg}) + \epsilon}$, measuring the target-to-background distinctness.
	\item \textbf{Intensity Fidelity}: $P_{ir} = \exp(-\frac{|\mu(\hat{x}_f^j | m_j) - \mu(x_{ir}^j | m_j)|}{\mu(x_{ir}^j | m_j) + \epsilon})$, enforcing absolute IR intensity preservation.
	\item \textbf{Signal-to-Noise Ratio}: $SNR = \frac{\mu(\hat{x}_f^j | \mathcal{M}_{in})}{\sigma(\hat{x}_f^j | \mathcal{M}_{bg}) + \epsilon}$, where $\sigma$ denotes the standard deviation.
\end{itemize}
The total prominence score is $\mathcal{S}_{tp} = 0.4 \cdot \sigma(2.0 \cdot C_w) + 0.3 \cdot P_{ir} + 0.3 \cdot \sigma(SNR - 3.0)$.

\paragraph{2. Local Sharpness ($\mathcal{S}_{ls}$)}
We compute the gradient energy to penalize over-smoothed. Applying a Laplacian operator $\mathbf{L}$ to the fused instance $\hat{x}_f^j$, we calculate the masked Tenengrad-variant sharpness:
\begin{equation}
	\mathcal{S}_{ls} = \text{Sigmoid}\left( \log \left( \frac{\sum_{p \in m_j} (\mathbf{L} * \hat{x}_f^j)^2}{\sum m_j + \epsilon} + 1 \right) - 3 \right).
\end{equation}

\paragraph{3. Information Entropy ($\mathcal{S}_{ie}$)}
This metric encourages high information density within the instance. Masked pixels are quantized into $L=256$ bins to compute Shannon entropy:
\begin{equation}
	H = -\sum_{i=0}^{L-1} p_i \log_2(p_i + \epsilon), \quad \mathcal{S}_{ie} = \frac{H}{8.0},
\end{equation}
where $p_i$ is the normalized intensity frequency within the mask $m_j$.

\paragraph{4. Infrared Visibility ($\mathcal{S}_{iv}$)}
To prevent the loss of IR-exclusive features (e.g., heat signatures), we compute the unique signal preservation ratio:
\begin{equation}
	\mathcal{R}_{ir} = \frac{\sum_{p \in m_j} \text{ReLU}(\hat{x}_f^j - x_{vis}^j)}{\sum_{p \in m_j} \text{ReLU}(x_{ir}^j - x_{vis}^j) + \epsilon}.
\end{equation}
This is averaged with the masked Pearson correlation between $\hat{x}_f^j$ and $x_{ir}^j$ and an intensity consistency score to evaluate structural and thermal alignment.

\paragraph{5. Structural Similarity ($\mathcal{S}_{ssim}$)}
We enforce consistency with both modalities using a masked SSIM index:
\begin{equation}
	\mathcal{S}_{ssim} = \frac{1}{2} \left[ \text{SSIM}(\hat{x}_f^j, x_{ir}^j | m_j) + \text{SSIM}(\hat{x}_f^j, x_{vis}^j | m_j) \right],
\end{equation}
where all internal statistics are calculated strictly over the support of $m_j$.

\paragraph{Reward Aggregation}
Individual metrics are aggregated into an instance score via a Softmax-weighted scheme $\mathbf{w}_{norm}^\top \mathbf{s}_j$. For an image with $N$ instances, the final batch reward is an area-weighted average to prioritize larger, semantically significant objects:
\begin{equation}
	\mathcal{J}_{inst} = \sum_{j=1}^{N} \frac{\text{Area}(m_j)}{\sum_{k=1}^N \text{Area}(m_k)} \cdot (\mathbf{w}_{norm}^\top \mathbf{s}_j).
\end{equation}

\section{VLM Reward Prompt Design}
\label{appendix:vlm_prompt}


To provide structured semantic feedback for Flow-GRPO, our prompt consists of a quality rubric and a structured assessment instruction. It guides the VLM to assess thermal target prominence, texture preservation, artifacts, and overall naturalness before assigning a numerical score.
Related work investigates fine-grained MLLM quality scoring with CoT~\cite{li2025nexttoken} providing context for our hybrid reward design.

\definecolor{promptgray}{RGB}{242, 242, 242}
\definecolor{promptborder}{RGB}{100, 100, 100}
\definecolor{tagred}{RGB}{220, 30, 30}

\begin{tcolorbox}[
	enhanced,
	attach boxed title to top left={yshift=-2mm, xshift=4mm},
	colback=promptgray,
	colframe=promptborder,
	colbacktitle=black,
	coltitle=white,
	title={System Prompt for IVIF Evaluation}, 
	fonttitle=\sffamily\bfseries\footnotesize,
	boxed title style={sharp corners, boxrule=0pt, left=1mm, right=1mm, top=0.5mm, bottom=0.5mm},
	arc=3pt,
	boxrule=0.5pt,
	width=\linewidth,
	fontupper=\ttfamily\footnotesize, 
	top=4mm, 
	left=3mm, right=3mm, bottom=3mm,
	before skip=10pt, after skip=10pt
	]
	\begin{flushleft} 
		Your role is to evaluate the perceptual quality and information fusion performance of Infrared and Visible Image Fusion (IVIF) results. \\
		1. \textbf{Bad}: The fusion fails significantly. Severe loss of information from either source (e.g., thermal targets are invisible or background texture is completely blurred). Significant artifacts (ghosting, severe noise, or distortion) make the image unusable. \\
		2. \textbf{Poor}: Unbalanced fusion. The image is dominated by one modality, causing a loss of complementary details. Thermal targets are dim, or visible textures are washed out. Noticeable artifacts like halo effects or unnatural edges are present. \\
		3. \textbf{Fair}: Acceptable fusion. Key thermal targets and basic background details are preserved. However, the contrast is low, the image lacks sharpness, or there is slight visual inconsistency (unnatural look) between the fused layers. \\
		4. \textbf{Good}: Effective fusion. High contrast with clear thermal targets and sharp background textures. Information from both sources is well-integrated with minimal artifacts. The image looks natural and informative. \\
		5. \textbf{Excellent}: Optimal synergy. Exceptional clarity and distinctness. Thermal targets are highly prominent, and background textures are rich and sharp. The fusion is seamless with zero artifacts, maximizing the complementary information for human or machine perception. \\
		Please first provide a detailed analysis of the evaluation process within the \textless{}Thought\textgreater{} tag. focus on: \\
		- \textbf{Thermal Target Prominence}: Are the infrared targets (e.g., pedestrians, cars) bright and distinct? \\
		- \textbf{Texture Detail Preservation}: Are the edges and background details from the visible image sharp? \\
		- \textbf{Artifacts}: Are there halo effects, ghosting, or noise introduced by the fusion? \\
		- \textbf{Overall Naturalness}: Is the overall brightness and contrast visually pleasing and natural? \\
		Then, give a final score from 1 to 5 within the \textless{}Score\textgreater{} tag. \\
		\textless{}Thought\textgreater{} \\
		{[}Analyze the evaluation process in detail here{]} \\
		\textless{}/Thought\textgreater{} \\
		\textless{}Score\textgreater{}X\textless{}/Score\textgreater{} \\
	\end{flushleft}
\end{tcolorbox}

\section{Extended Experimental Results}
Due to space limitations in the main text, we present additional qualitative and quantitative evaluations for both clean and degraded scenarios here. While existing SOTA methods exhibit a pronounced decline in performance under severe environmental perturbations, our method maintains exceptional fusion quality across all conditions. These results further validate the robustness and versatility of our framework in both ideal and challenging real-world environments.

\begin{figure*}[]
	\centering
	\includegraphics[width=0.98\linewidth]{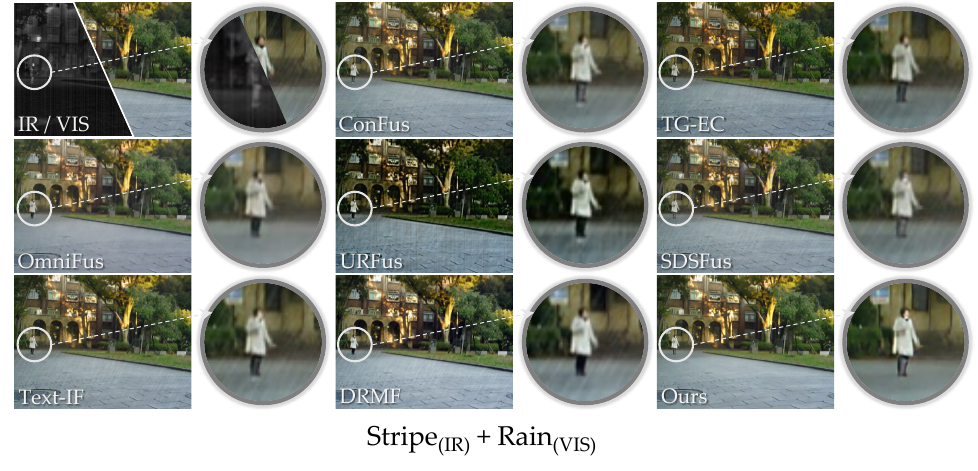}
	\caption{Fusion qualitative comparisons between SOTA methods and our OneHRL. It presents stripe (infrared images) with rain (visible images) degraded scenarios in the MSRS dataset. In all cases, OneHRL achieves well-performance results.}
	\label{fig:appendix-msrs-2}	
\end{figure*}
\begin{figure*}[]
	\centering
	\includegraphics[width=0.98\linewidth]{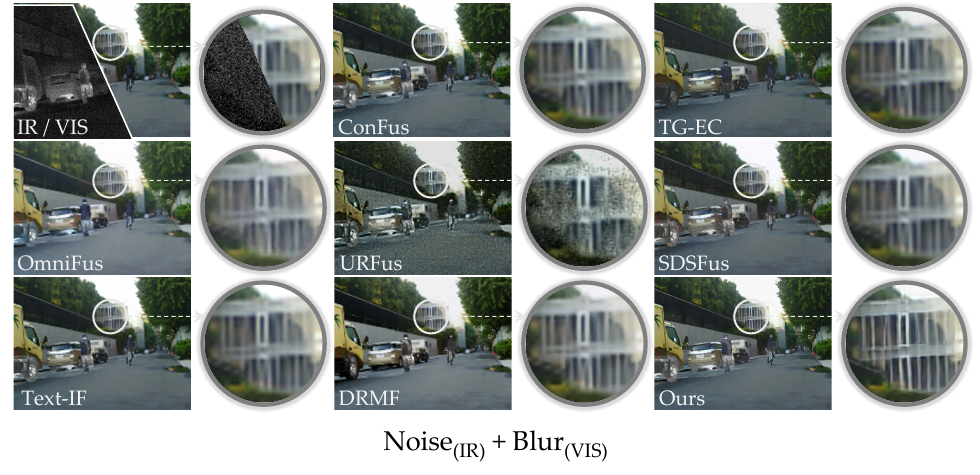}
	\caption{Fusion qualitative comparisons between SOTA methods and our OneHRL. It illustrates noise (infrared images) with blur (visible images) degraded scenarios in the MSRS dataset. In all cases, OneHRL achieves well-performance results.}
	\label{fig:appendix-msrs-1}	
\end{figure*}

\begin{table}[]
	\centering
	\caption{Quantitative comparison of fusion under Noise(IR)-Blur(VIS) scenario.}
	\begin{tabular}{c|ccccc|ccccc}
		\toprule
		\multirow{2}{*}{\textbf{Method}} 
		& \multicolumn{5}{c|}{\textbf{$\text{M}^\text{3}$FD}} 
		& \multicolumn{5}{c}{\textbf{MSRS}} 
		\\
		& CC & SCD & MI & VIF & MSE
		& CC & SCD & MI & VIF & MSE
		\\ 
		\midrule
		ConFus  & 0.473 & 1.342 & 2.748 & \secondhl{\textbf{0.400}} & 0.059 & 0.576 & 1.421 & 2.859 & 0.617 & 0.043
		\\
		DRMF    & 0.414 & 1.006 & \besthl{\textbf{3.360}} & 0.384 & 0.067 & 0.509 & 1.051 & \secondhl{\textbf{3.243}} & \secondhl{\textbf{0.635}} & 0.068
		\\
		OmniFus & 0.443 & 1.188 & 3.011 & 0.393 & 0.060 & 0.529 & 1.114 & 2.640 & 0.605 & 0.064
		\\
		SDSFus  & 0.432 & 1.165 & 2.861 & 0.300 & 0.059 & 0.577 & 1.333 & 2.265 & 0.439 & 0.039
		\\
		Text-IF & 0.387 & 1.099 & 1.050 & 0.200 & 0.056 & \secondhl{\textbf{0.586}} & \secondhl{\textbf{1.444}} & 2.655 & 0.555 & 0.040
		\\
		TG-EC   & 0.458 & 1.233 & 2.718 & 0.370 & 0.051 & 0.565 & 1.148 & 2.413 & 0.464 & \secondhl{\textbf{0.034}}
		\\
		URFus   & \secondhl{\textbf{0.500}} & \secondhl{\textbf{1.498}} & 1.278 & 0.237 & \secondhl{\textbf{0.050}} & 0.581 & 1.361 & 1.418 & 0.360 & 0.050
		\\
		Ours    & \besthl{\textbf{0.511}} & \besthl{\textbf{1.506}} & \secondhl{\textbf{3.116}} & \besthl{\textbf{0.452}} & \besthl{\textbf{0.049}} & \besthl{\textbf{0.614}} & \besthl{\textbf{1.549}} & \besthl{\textbf{3.713}} & \besthl{\textbf{0.673}} & \besthl{\textbf{0.032}}
		\\
		\bottomrule
	\end{tabular}
\end{table}

\begin{figure*}[]
	\centering
	\includegraphics[width=0.98\linewidth]{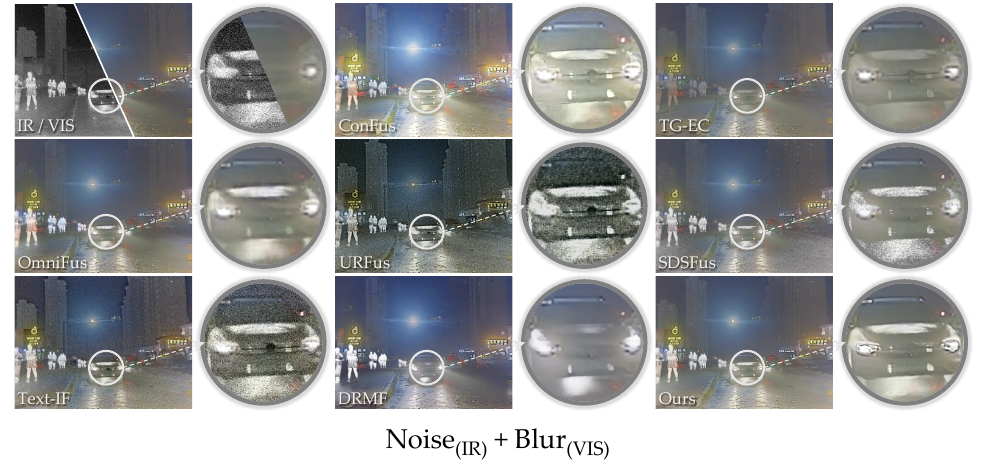}
	\caption{Fusion qualitative comparisons between SOTA methods and our OneHRL. It illustrates noise (infrared images) with blur (visible images) degraded scenarios in the $\text{M}^3\text{FD}$ dataset. In all cases, OneHRL achieves well-performance results.}
	\label{fig:appendix-m3fd-1}	
\end{figure*}
\begin{figure*}[]
	\centering
	\includegraphics[width=0.98\linewidth]{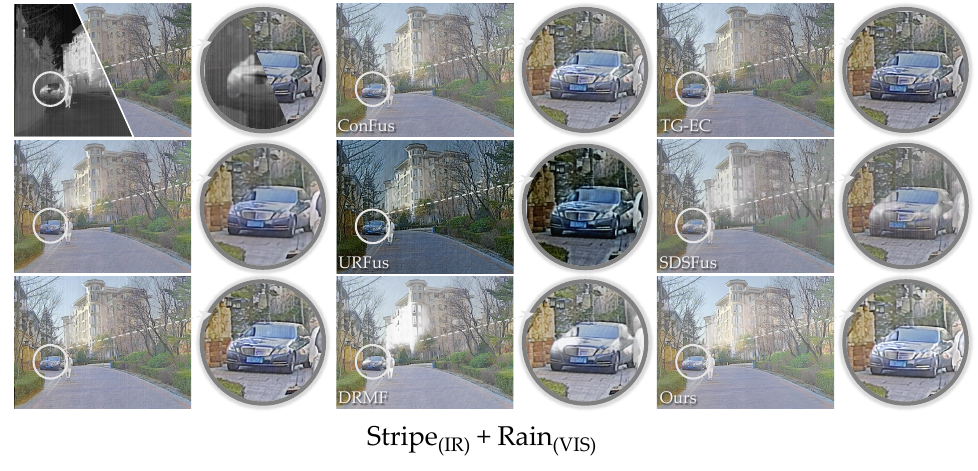}
	\caption{Fusion qualitative comparisons between SOTA methods and our OneHRL. It presents stripe (infrared images) with rain (visible images) degraded scenarios in the $\text{M}^3\text{FD}$ dataset. In allcases, OneHRL achieves well-performance results.}
	\label{fig:appendix-m3fd-2}	
\end{figure*}

\begin{table}[]
	\centering
	\caption{Quantitative comparison of fusion under Stripe(IR)-Rain(VIS) scenario.}
	\begin{tabular}{c|ccccc|ccccc}
		\toprule
		\multirow{2}{*}{\textbf{Method}} 
		& \multicolumn{5}{c|}{\textbf{$\text{M}^\text{3}$FD}} 
		& \multicolumn{5}{c}{\textbf{MSRS}} 
		\\
		& CC & SCD & MI & VIF & MSE
		& CC & SCD & MI & VIF & MSE
		\\ 
		\midrule
		ConFus  & 0.488 & 1.381 & 2.851 & 0.621 & 0.056 & \secondhl{\textbf{0.600}} & \secondhl{\textbf{1.529}} & 2.952 & \secondhl{\textbf{0.757}} & 0.035
		\\
		DRMF    & 0.431 & 1.081 & \besthl{\textbf{3.331}} & \secondhl{\textbf{0.632}} & 0.070 & 0.519 & 1.103 & {\textbf{3.114}} & 0.678 & 0.064
		\\
		OmniFus & 0.472 & 1.294 & 2.839 & 0.555 & 0.051 & 0.535 & 1.105 & 2.506 & 0.638 & 0.062
		\\
		SDSFus  & 0.475 & 1.304 & 3.280 & 0.572 & 0.049 & 0.597 & 1.438 & 2.879 & 0.672 & 0.033
		\\
		Text-IF & 0.481 & 1.394 & 2.552 & 0.595 & 0.056 & 0.589 & 1.490 & 3.049 & 0.728 & 0.035
		\\
		TG-EC   & 0.502 & 1.484 & 2.582 & 0.603 & \secondhl{\textbf{0.045}} & 0.578 & 1.326 & 2.480 & 0.603 & \besthl{\textbf{0.029}}
		\\
		URFus   & \secondhl{\textbf{0.538}} & \secondhl{\textbf{1.617}} & 1.998 & 0.553 & 0.048 & 0.587 & 1.414 & 1.919 & 0.626 & 0.050
		\\
		Ours    & \besthl{\textbf{0.543}} & \besthl{\textbf{1.644}} & \secondhl{\textbf{3.310}} & \besthl{\textbf{0.673}} & \besthl{\textbf{0.045}} & \besthl{\textbf{0.620}} & \besthl{\textbf{1.618}} & \besthl{\textbf{3.382}} & \besthl{\textbf{0.775}} & \secondhl{\textbf{0.032}}
		\\
		\bottomrule
	\end{tabular}
\end{table}

\begin{table}[]
	\centering
	\caption{Quantitative comparison of detection under no degradation scenario.}
	\begin{tabular}{c|cccccc|c}
		\toprule
		&\multicolumn{6}{c|}{$\mathrm{AP@.5}$} & 
		\\
		\multirow{-2}{*}{\textbf{Method}} & People & Car & Bus & Motor & Truck & Lamp &
		\multirow{-2}{*}{$\mathrm{mAP@.5}$}     
		\\ 
		\midrule
		ConFus & 0.655 & \secondhl{\textbf{0.866}} & 0.733 & 0.449 & 0.692 & \secondhl{\textbf{0.533}} & 0.655 
		\\
		DRMF & 0.590 & 0.831 & 0.683 & 0.342 & 0.563 & 0.458 & 0.578                                 
		\\
		OmniFus & 0.616 & 0.843 & 0.690 & 0.265 & 0.679 & 0.506 & 0.600                                 
		\\
		SDSFus & 0.612 & 0.769 & 0.505 & 0.289 & 0.449 & 0.288 & 0.486                                 
		\\
		Text-IF & \besthl{\textbf{0.658}} & 0.858 & 0.751 & 0.468 & 0.701 & 0.522 & 0.659                                
		\\
		TG-EC & 0.614 & 0.856 & \secondhl{\textbf{0.796}} &\secondhl {\textbf{0.507}} & \secondhl{\textbf{0.715}} & 0.526 & \secondhl{\textbf{0.669}}                                 
		\\
		URFus & 0.610 & 0.850 & 0.779 & 0.426 & 0.702 & 0.513 & 0.646                                 
		\\
		Ours & \secondhl{\textbf{0.656}} & \besthl{\textbf{0.872}} & \besthl{\textbf{0.803}} & \besthl{\textbf{0.512}} & \besthl{\textbf{0.718}} & \besthl{\textbf{0.538}} & \besthl{\textbf{0.683}}
		\\ 
		\bottomrule
	\end{tabular}
\end{table}

\begin{table}[]
	\centering
	\caption{Quantitative comparison of segmentation under no degradation scenario.}
	\renewcommand\tabcolsep{6pt} 
	\begin{tabular}{c|ccccccccc|c}
		\toprule
		&\multicolumn{9}{c|}{$\mathrm{IoU}$} & 
		\\
		\multirow{-2}{*}{\textbf{Method}} & BG & Car & Person & Bike & Curve & C S & GR & C C & Bump &
		\multirow{-2}{*}{$\mathrm{mIoU}$}     
		\\ 
		\midrule
		ConFus & 0.973 & 0.827 & \secondhl{\textbf{0.604}} & 0.616 & \secondhl{\textbf{0.224}} & 0.477 & 0.571 & 0.327 & \secondhl{\textbf{0.481}} & 0.567
		\\
		DRMF & 0.970 & 0.795 & 0.564 & 0.601 & 0.133 & 0.489 & \secondhl{\textbf{0.638}} & 0.424 & 0.324 & 0.549                                
		\\
		OmniFus & 0.970 & 0.803 & 0.521 & 0.556 & 0.121 & 0.483 & 0.495 & 0.416 & 0.365 & 0.526                                 
		\\
		SDSFus & 0.972 & 0.813 & 0.584 & 0.582 & 0.161 & 0.531 & 0.587 & 0.371 & 0.433 & 0.560                                 
		\\
		Text-IF & \secondhl{\textbf{0.973}} & \secondhl{\textbf{0.828}} & 0.587 & \secondhl{\textbf{0.625}} & 0.127 & \secondhl{\textbf{0.535}} & 0.607 & \secondhl{\textbf{0.426}} & 0.425 & \secondhl{\textbf{0.570}}                              
		\\
		TG-EC & 0.971 & 0.793 & 0.594 & 0.590 & 0.098 & 0.493 & 0.553 & 0.358 & 0.404 & 0.539                                 
		\\
		URFus & 0.971 & 0.813 & 0.579 & 0.597 & 0.163 & 0.510 & 0.543 & 0.358 & 0.430 & 0.552                                 
		\\
		Ours & \besthl{\textbf{0.976}} & \besthl{\textbf{0.842}} & \besthl{\textbf{0.623}} & \besthl{\textbf{0.626}} & \besthl{\textbf{0.271}} & \besthl{\textbf{0.590}} & \besthl{\textbf{0.699}} & \besthl{\textbf{0.445}} & \besthl{\textbf{0.502}} & \besthl{\textbf{0.619}}
		\\ 
		\bottomrule
	\end{tabular}
\end{table}

\begin{table}[]
	\centering
	\caption{Quantitative comparison of detection under Noise(IR)-Blur(VIS) scenario.}
	\begin{tabular}{c|cccccc|c}
		\toprule
		&\multicolumn{6}{c|}{$\mathrm{AP@.5}$} & 
		\\
		\multirow{-2}{*}{\textbf{Method}} & People & Car & Bus & Motor & Truck & Lamp &
		\multirow{-2}{*}{$\mathrm{mAP@.5}$}     
		\\ 
		\midrule
		ConFus & \secondhl{\textbf{0.621}} & 0.782 & 0.680 & 0.310 & 0.548 & 0.356 & 0.550 
		\\
		DRMF & 0.514 & 0.716 & 0.486 & 0.269 & 0.326 & 0.213 & 0.421                                 
		\\
		OmniFus & 0.600 & \secondhl{\textbf{0.814}} & 0.627 & 0.226 & \secondhl{\textbf{0.589}} & 0.398 & 0.542                                 
		\\
		SDSFus & 0.457 & 0.549 & 0.103 & 0.065 & 0.132 & 0.162 & 0.245                                 
		\\
		Text-IF & 0.440 & 0.406 & 0.053 & 0.002 & 0.011 & 0.039 & 0.159                                
		\\
		TG-EC & 0.532 & 0.773 & \secondhl{\textbf{0.724}} & \secondhl{\textbf{0.380}} & 0.551 & \secondhl{\textbf{0.412}} & \secondhl{\textbf{0.562}}                                 
		\\
		URFus & 0.501 & 0.467 & 0.275 & 0.045 & 0.097 & 0.094 & 0.247                                 
		\\
		Ours & \besthl{\textbf{0.629}} & \besthl{\textbf{0.798}} & \besthl{\textbf{0.756}} & \besthl{\textbf{0.394}} & \besthl{\textbf{0.652}} & \besthl{\textbf{0.497}} & \besthl{\textbf{0.621}} 
		\\ 
		\bottomrule
	\end{tabular}
\end{table}

\begin{table}[]
	\centering
	\caption{Quantitative comparison of segmentation under Noise(IR)-Blur(VIS) scenario.}
	\renewcommand\tabcolsep{6pt} 
	\begin{tabular}{c|ccccccccc|c}
		\toprule
		&\multicolumn{9}{c|}{$\mathrm{IoU}$} & 
		\\
		\multirow{-2}{*}{\textbf{Method}} & BG & Car & Person & Bike & Curve & C S & GR & C C & Bump &
		\multirow{-2}{*}{$\mathrm{mIoU}$}     
		\\ 
		\midrule
		ConFus & \secondhl{\textbf{0.970}} & \secondhl{\textbf{0.797}} & \secondhl{\textbf{0.556}} & \secondhl{\textbf{0.542}} & 0.173 & 0.478 & 0.449 & 0.335 & \secondhl{\textbf{0.444}} & \secondhl{\textbf{0.527}}
		\\
		DRMF & 0.968 & 0.771 & 0.533 & 0.532 & \secondhl{\textbf{0.186}} & \secondhl{\textbf{0.482}} & 0.511 & 0.374 & 0.332 & 0.521                                
		\\
		OmniFus & 0.967 & 0.757 & 0.484 & 0.499 & 0.142 & 0.443 & 0.348 & \secondhl{\textbf{0.382}} & 0.441 & 0.496                                 
		\\
		SDSFus & 0.959 & 0.659 & 0.315 & 0.279 & 0.008 & 0.305 & 0.423 & 0.234 & 0.402 & 0.407                                 
		\\
		Text-IF & 0.968 & 0.775 & 0.541 & 0.505 & 0.140 & 0.461 & 0.497 & 0.323 & 0.412 & 0.513                              
		\\
		TG-EC & 0.968 & 0.778 & 0.486 & 0.522 & 0.105 & 0.444 & \secondhl{\textbf{0.541}} & 0.328 & 0.403 & 0.508                                 
		\\
		URFus & 0.950 & 0.504 & 0.315 & 0.251 & 0.047 & 0.194 & 0.005 & 0.165 & 0.001 & 0.275                                 
		\\
		Ours & \besthl{\textbf{0.974}} & \besthl{\textbf{0.824}} & \besthl{\textbf{0.597}} & \besthl{\textbf{0.569}} & \besthl{\textbf{0.249}} & \besthl{\textbf{0.589}} & \besthl{\textbf{0.687}} & \besthl{\textbf{0.436}} & \besthl{\textbf{0.510}} & \besthl{\textbf{0.601}}
		\\ 
		\bottomrule
	\end{tabular}
\end{table}

\begin{table}[]
	\centering
	\caption{Quantitative comparison of detection under Stripe(IR)-Rain(VIS) scenario.}
	\begin{tabular}{c|cccccc|c}
		\toprule
		&\multicolumn{6}{c|}{$\mathrm{AP@.5}$} & 
		\\
		\multirow{-2}{*}{\textbf{Method}} & People & Car & Bus & Motor & Truck & Lamp &
		\multirow{-2}{*}{$\mathrm{mAP@.5}$}     
		\\ 
		\midrule
		ConFus & \secondhl{\textbf{0.653}} & \secondhl{\textbf{0.851}} & 0.717 & \secondhl{\textbf{0.414}} & 0.635 & 0.495 & 0.628 
		\\
		DRMF & 0.582 & 0.786 & 0.436 & 0.264 & 0.506 & 0.393 & 0.494                                 
		\\
		OmniFus & 0.598 & 0.811 & 0.629 & 0.238 & 0.593 & 0.406 & 0.546                                 
		\\
		SDSFus & 0.605 & 0.682 & 0.261 & 0.138 & 0.251 & 0.215 & 0.359                                 
		\\
		Text-IF & \besthl{\textbf{0.653}} & 0.796 & 0.641 & 0.296 & 0.593 & 0.463 & 0.574                               
		\\
		TG-EC & 0.563 & 0.801 & 0.660 & 0.353 & 0.575 & 0.483 & 0.572                                 
		\\
		URFus & 0.609 & 0.791 & \secondhl{\textbf{0.747}} & 0.391 & \besthl{\textbf{0.676}} & \besthl{\textbf{0.580}} & \secondhl{\textbf{0.632}}                                 
		\\
		Ours & 0.651 & \besthl{\textbf{0.863}} & \besthl{\textbf{0.759}} & \besthl{\textbf{0.418}} & \secondhl{\textbf{0.656}} & \secondhl{\textbf{0.518}} & \besthl{\textbf{0.644}} 
		\\ 
		\bottomrule
	\end{tabular}
\end{table}

\begin{table}[]
	\centering
	\caption{Quantitative comparison of segmentation under Stripe(IR)-Rain(VIS) scenario.}
	\renewcommand\tabcolsep{6pt} 
	\begin{tabular}{c|ccccccccc|c}
		\toprule
		&\multicolumn{9}{c|}{$\mathrm{IoU}$} & 
		\\
		\multirow{-2}{*}{\textbf{Method}} & BG & Car & Person & Bike & Curve & C S & GR & C C & Bump &
		\multirow{-2}{*}{$\mathrm{mIoU}$}     
		\\ 
		\midrule
		ConFus & \secondhl{\textbf{0.965}} & 0.711 & \secondhl{\textbf{0.545}} & \secondhl{\textbf{0.563}} & 0.001 & \secondhl{\textbf{0.392}} & 0.306 & 0.326 & 0.143 & 0.441
		\\
		DRMF & 0.961 & 0.681 & 0.490 & 0.425 & 0.003 & 0.230 & 0.264 & 0.290 & 0.008 & 0.385                                
		\\
		OmniFus & 0.965 & \secondhl{\textbf{0.756}} & 0.466 & 0.489 & \secondhl{\textbf{0.007}} & 0.317 & \secondhl{\textbf{0.326}} & \secondhl{\textbf{0.357}} & \secondhl{\textbf{0.289}} & \secondhl{\textbf{0.449}}                                 
		\\
		SDSFus & 0.958 & 0.585 & 0.460 & 0.460 & 0.001 & 0.289 & 0.236 & 0.249 & 0.115 & 0.374                                 
		\\
		Text-IF & 0.961 & 0.661 & 0.511 & 0.484 & 0.001 & 0.230 & 0.283 & 0.264 & 0.106 & 0.390                              
		\\
		TG-EC & 0.956 & 0.596 & 0.361 & 0.439 & 0.001 & 0.208 & 0.243 & 0.220 & 0.139 & 0.352                                 
		\\
		URFus & 0.962 & 0.685 & 0.513 & 0.498 & 0.003 & 0.303 & 0.297 & 0.173 & 0.142 & 0.401                                 
		\\
		Ours & \besthl{\textbf{0.975}} & \besthl{\textbf{0.832}} & \besthl{\textbf{0.617}} & \besthl{\textbf{0.609}} & \besthl{\textbf{0.253}} & \besthl{\textbf{0.566}} & \besthl{\textbf{0.653}} & \besthl{\textbf{0.447}} & \besthl{\textbf{0.530}} & \besthl{\textbf{0.609}}
		\\ 
		\bottomrule
	\end{tabular}
\end{table}

\begin{table}[]
	\centering
	\caption{Quantitative comparison of detection under NoiseStripe(IR)-HazeNoise(VIS) scenario.}
	\begin{tabular}{c|cccccc|c}
		\toprule
		&\multicolumn{6}{c|}{$\mathrm{AP@.5}$} & 
		\\
		\multirow{-2}{*}{\textbf{Method}} & People & Car & Bus & Motor & Truck & Lamp &
		\multirow{-2}{*}{$\mathrm{mAP@.5}$}     
		\\ 
		\midrule
		ConFus & \secondhl{\textbf{0.522}} & \secondhl{\textbf{0.475}} & 0.094 & 0.010 & 0.041 & 0.030 & \secondhl{\textbf{0.196}} 
		\\
		DRMF & 0.431 & 0.400 & \secondhl{\textbf{0.101}} & \secondhl{\textbf{0.026}} & 0.034 & / & 0.165                                 
		\\
		OmniFus & 0.350 & 0.467 & 0.022 & 0.011 & \secondhl{\textbf{0.069}} & 0.041 & 0.160                                 
		\\
		SDSFus & 0.249 & 0.230 & / & / & 0.011 & / & 0.081                                 
		\\
		Text-IF & 0.365 & 0.411 & 0.091 & / & 0.016 & \secondhl{\textbf{0.054}} & 0.156                              
		\\
		TG-EC & 0.316 & 0.397 & 0.053 & / & 0.023 & 0.012 & 0.134                                 
		\\
		URFus & 0.477 & 0.447 & 0.048 & / & 0.036 & 0.022 & 0.172                                 
		\\
		Ours & \besthl{\textbf{0.606}} & \besthl{\textbf{0.760}} & \besthl{\textbf{0.572}} & \besthl{\textbf{0.161}} & \besthl{\textbf{0.490}} & \besthl{\textbf{0.228}} & \besthl{\textbf{0.470}} 
		\\ 
		\bottomrule
	\end{tabular}
\end{table}

\begin{table}[]
	\centering
	\caption{Quantitative comparison of segmentation under NoiseStripe(IR)-HazeNoise(VIS) scenario.}
	\renewcommand\tabcolsep{6pt} 
	\begin{tabular}{c|ccccccccc|c}
		\toprule
		&\multicolumn{9}{c|}{$\mathrm{IoU}$} & 
		\\
		\multirow{-2}{*}{\textbf{Method}} & BG & Car & Person & Bike & Curve & C S & GR & C C & Bump &
		\multirow{-2}{*}{$\mathrm{mIoU}$}     
		\\ 
		\midrule
		ConFus & 0.942 & 0.363 & 0.216 & \secondhl{\textbf{0.279}} & 0.001 & 0.005 & / & 0.004 & / & 0.213
		\\
		DRMF & 0.931 & 0.007 & \secondhl{\textbf{0.347}} & 0.007 & / & / & / & / & / & 0.159                                
		\\
		OmniFus & \secondhl{\textbf{0.945}} & \secondhl{\textbf{0.440}} & 0.272 & 0.185 & / & \secondhl{\textbf{0.008}} & / & \secondhl{\textbf{0.109}} & \secondhl{\textbf{0.008}} & \secondhl{\textbf{0.236}}                                 
		\\
		SDSFus & 0.927 & 0.007 & / & 0.006 & / & / & / & / & / & 0.119                                 
		\\
		Text-IF & 0.928 & 0.006 & 0.003 & 0.007 & / & / & / & / & / & 0.123                              
		\\
		TG-EC & 0.944 & 0.417 & 0.232 & 0.171 & / & 0.007 & \secondhl{\textbf{0.001}} & 0.008 & 0.007 & 0.223                                 
		\\
		URFus & 0.933 & 0.159 & 0.185 & 0.212 & 0.001 & \secondhl{\textbf{0.001}} & / & / & / & 0.169                                 
		\\
		Ours & \besthl{\textbf{0.970}} & \besthl{\textbf{0.772}} & \besthl{\textbf{0.575}} & \besthl{\textbf{0.459}} & \besthl{\textbf{0.222}} & \besthl{\textbf{0.484}} & \besthl{\textbf{0.509}} & \besthl{\textbf{0.369}} & \besthl{\textbf{0.390}} & \besthl{\textbf{0.528}}
		\\ 
		\bottomrule
	\end{tabular}
\end{table}

\begin{table}[]
	\centering
	\caption{Quantitative comparison of detection under NoiseStripe(IR)-HazeRain(VIS) scenario.}
	\begin{tabular}{c|cccccc|c}
		\toprule
		&\multicolumn{6}{c|}{$\mathrm{AP@.5}$} & 
		\\
		\multirow{-2}{*}{\textbf{Method}} & People & Car & Bus & Motor & Truck & Lamp &
		\multirow{-2}{*}{$\mathrm{mAP@.5}$}     
		\\ 
		\midrule
		ConFus & \secondhl{\textbf{0.621}} & \secondhl{\textbf{0.762}} & \secondhl{\textbf{0.535}} & \secondhl{\textbf{0.222}} & 0.305 & \secondhl{\textbf{0.299}} & \secondhl{\textbf{0.457}} 
		\\
		DRMF & 0.535 & 0.667 & 0.330 & 0.170 & \secondhl{\textbf{0.333}} & 0.166 & 0.367                                 
		\\
		OmniFus & 0.393 & 0.582 & 0.238 & 0.026 & 0.167 & 0.089 & 0.249                                 
		\\
		SDSFus & 0.283 & 0.552 & 0.038 & 0.008 & 0.065 & 0.101 & 0.175                                 
		\\
		Text-IF & 0.444 & 0.621 & 0.251 & 0.096 & 0.142 & 0.158 & 0.285                              
		\\
		TG-EC & 0.202 & 0.580 & 0.219 & 0.041 & 0.194 & 0.150 & 0.231                                 
		\\
		URFus & 0.469 & 0.400 & 0.024 & / & 0.005 & 0.019 & 0.153                                 
		\\
		Ours & \besthl{\textbf{0.636}} & \besthl{\textbf{0.820}} & \besthl{\textbf{0.623}} & \besthl{\textbf{0.330}} & \besthl{\textbf{0.599}} & \besthl{\textbf{0.400}} & \besthl{\textbf{0.568}} 
		\\ 
		\bottomrule
	\end{tabular}
\end{table}

\begin{table}[h]
	\centering
	\caption{Quantitative comparison of segmentation under NoiseStripe(IR)-HazeRain(VIS) scenario.}
	\renewcommand\tabcolsep{6pt} 
	\begin{tabular}{c|ccccccccc|c}
		\toprule
		&\multicolumn{9}{c|}{$\mathrm{IoU}$} & 
		\\
		\multirow{-2}{*}{\textbf{Method}} & BG & Car & Person & Bike & Curve & C S & GR & C C & Bump &
		\multirow{-2}{*}{$\mathrm{mIoU}$}     
		\\ 
		\midrule
		ConFus & \secondhl{\textbf{0.964}} & \secondhl{\textbf{0.696}} & \secondhl{\textbf{0.488}} & \secondhl{\textbf{0.542}} & \secondhl{\textbf{0.001}} & \secondhl{\textbf{0.319}} & 0.008 & \secondhl{\textbf{0.287}} & \secondhl{\textbf{0.182}} & \secondhl{\textbf{0.397}}
		\\
		DRMF & 0.954 & 0.557 & 0.438 & 0.337 & / & 0.110 & 0.108 & 0.174 & 0.004 & 0.304                                
		\\
		OmniFus & 0.949 & 0.509 & 0.274 & 0.247 & / & 0.134 & \secondhl{\textbf{0.135}} & 0.008 & 0.105 & 0.271                                 
		\\
		SDSFus & 0.938 & 0.280 & 0.129 & 0.219 & / & 0.005 & 0.003 & 0.007 & / & 0.193                                
		\\
		Text-IF & 0.945 & 0.387 & 0.262 & 0.283 & / & 0.006 & 0.002 & 0.106 & 0.003 & 0.233                              
		\\
		TG-EC & 0.940 & 0.311 & 0.413 & 0.243 & / & 0.005 & 0.001 & 0.008 & / & 0.199                                 
		\\
		URFus & 0.933 & 0.134 & 0.212 & 0.240 & / & 0.004 & / & / & / & 0.174                                 
		\\
		Ours & \besthl{\textbf{0.974}} & \besthl{\textbf{0.818}} & \besthl{\textbf{0.602}} & \besthl{\textbf{0.566}} & \besthl{\textbf{0.224}} & \besthl{\textbf{0.572}} & \besthl{\textbf{0.636}} & \besthl{\textbf{0.441}} & \besthl{\textbf{0.483}} & \besthl{\textbf{0.591}}
		\\ 
		\bottomrule
	\end{tabular}
\end{table}


\newpage
\clearpage

\section*{NeurIPS Paper Checklist}

\begin{enumerate}

\item {\bf Claims}
    \item[] Question: Do the main claims made in the abstract and introduction accurately reflect the paper's contributions and scope?
    \item[] Answer: \answerYes{} 
    \item[] Justification: The abstract and introduction of the paper clearly state the claims made, including the contributions made in the paper and important assumptions and limitations.

\item {\bf Limitations}
    \item[] Question: Does the paper discuss the limitations of the work performed by the authors?
    \item[] Answer: \answerNo{} 
    \item[] Justification: The paper do not discusses the limitations of the work.

\item {\bf Theory assumptions and proofs}
    \item[] Question: For each theoretical result, does the paper provide the full set of assumptions and a complete (and correct) proof?
    \item[] Answer: \answerYes{} 
    \item[] Justification: All the theorems, formulas, and proofs in the paper have been numbered and cross-referenced, and all assumptions have been clearly stated or referenced in the statement of any theorems.

    \item {\bf Experimental result reproducibility}
    \item[] Question: Does the paper fully disclose all the information needed to reproduce the main experimental results of the paper to the extent that it affects the main claims and/or conclusions of the paper (regardless of whether the code and data are provided or not)?
    \item[] Answer: \answerYes{} 
    \item[] Justification: The paper has fully disclosed all the information needed to reproduce the main experimental results of the paper to the extent that it affects the main claims and conclusions of the paper.

\item {\bf Open access to data and code}
    \item[] Question: Does the paper provide open access to the data and code, with sufficient instructions to faithfully reproduce the main experimental results, as described in supplemental material?
    \item[] Answer: \answerYes{} 
    \item[] Justification: We will open access to the code after the paper is accepted.

\item {\bf Experimental setting/details}
    \item[] Question: Does the paper specify all the training and test details (e.g., data splits, hyperparameters, how they were chosen, type of optimizer) necessary to understand the results?
    \item[] Answer: \answerYes{} 
    \item[] Justification: We have specified all the training and test details necessary to understand the results.

\item {\bf Experiment statistical significance}
    \item[] Question: Does the paper report error bars suitably and correctly defined or other appropriate information about the statistical significance of the experiments?
    \item[] Answer: \answerNo{} 
    \item[] Justification: The paper does not report error bars.

\item {\bf Experiments compute resources}
    \item[] Question: For each experiment, does the paper provide sufficient information on the computer resources (type of compute workers, memory, time of execution) needed to reproduce the experiments?
    \item[] Answer: \answerYes{} 
    \item[] Justification: The paper provides sufficient information on the computer resource for experiments.
    
\item {\bf Code of ethics}
    \item[] Question: Does the research conducted in the paper conform, in every respect, with the NeurIPS Code of Ethics \url{https://neurips.cc/public/EthicsGuidelines}?
    \item[] Answer: \answerYes{} 
    \item[] Justification: The research conducted in the paper conforms in every respect with the NeurIPS Code of Ethics, ensuring ethical standards are upheld throughout the study.

\item {\bf Broader impacts}
    \item[] Question: Does the paper discuss both potential positive societal impacts and negative societal impacts of the work performed?
    \item[] Answer: \answerYes{} 
    \item[] Justification: The paper discusses both potential positive societal impacts and negative societal impacts of the work performed.
    
\item {\bf Safeguards}
    \item[] Question: Does the paper describe safeguards that have been put in place for responsible release of data or models that have a high risk for misuse (e.g., pre-trained language models, image generators, or scraped datasets)?
    \item[] Answer: \answerYes{} 
    \item[] Justification:  The paper describes safeguards that have been put in place for the responsible release of data or models that have a high risk for misuse, ensuring that appropriate measures are taken to mitigate potential risks.

\item {\bf Licenses for existing assets}
    \item[] Question: Are the creators or original owners of assets (e.g., code, data, models), used in the paper, properly credited and are the license and terms of use explicitly mentioned and properly respected?
    \item[] Answer: \answerYes{} 
    \item[] Justification: The creators or original owners of assets used in the paper are properly credited, and the license and terms of use are explicitly mentioned and properly respected.

\item {\bf New assets}
    \item[] Question: Are new assets introduced in the paper well documented and is the documentation provided alongside the assets?
    \item[] Answer: \answerYes{} 
    \item[] Justification: The new assets introduced in the paper are well documented, and the documentation is provided alongside the assets.

\item {\bf Crowdsourcing and research with human subjects}
    \item[] Question: For crowdsourcing experiments and research with human subjects, does the paper include the full text of instructions given to participants and screenshots, if applicable, as well as details about compensation (if any)? 
    \item[] Answer: \answerNA{} 
    \item[] Justification: The paper does not involve crowdsourcing nor research with human subjects.

\item {\bf Institutional review board (IRB) approvals or equivalent for research with human subjects}
    \item[] Question: Does the paper describe potential risks incurred by study participants, whether such risks were disclosed to the subjects, and whether Institutional Review Board (IRB) approvals (or an equivalent approval/review based on the requirements of your country or institution) were obtained?
    \item[] Answer: \answerNA{} 
    \item[] Justification: The paper does not involve crowdsourcing nor research with human subjects.

\item {\bf Declaration of LLM usage}
    \item[] Question: Does the paper describe the usage of LLMs if it is an important, original, or non-standard component of the core methods in this research? Note that if the LLM is used only for writing, editing, or formatting purposes and does \emph{not} impact the core methodology, scientific rigor, or originality of the research, declaration is not required.
    \item[] Answer: \answerNA{} 
    \item[] Justification: The core method development in the paper does not involve LLMs as any important, original, or non-standard components.

\end{enumerate}

\end{document}